\pdfoutput=1
\documentclass{article}
\usepackage{times,authblk}
\usepackage{fullpage}

\usepackage{geometry}
\usepackage{hyperref}
\usepackage{rotating, color,subfigure,algorithm,algorithmic}

\usepackage{multirow}
\usepackage{graphicx}

\usepackage{enumerate}
\usepackage{amsthm,amsmath,amssymb}
\usepackage{comment}

\usepackage{mkolar_definitions}

\newcount\Comments
\usepackage{ifthen}
\definecolor{darkgreen}{rgb}{0,0.5,0}
\definecolor{darkred}{rgb}{0.7,0,0}
\definecolor{teal}{rgb}{0.3,0.8,0.8}
\newcommand{\kibitz}[2]{\ifnum\Comments=1\textcolor{#1}{#2}\fi}

\newcommand{\numcolumns}{one}
\newcommand{\version}{arxiv}
\ifthenelse{\equal{\version}{arxiv}}{
\newcommand{\figsizeb}{0.18}
\newcommand{\figsizea}{0.25}
}{
\newcommand{\figsizeb}{0.21}
\newcommand{\figsizea}{0.25}
}
\begin{document}
\title{Extreme Compressive Sampling for Covariance Estimation}

\author[1]{
Martin Azizyan
\thanks{mazizyan@cs.cmu.edu}}
\author[2]{
Akshay Krishnamurthy
\thanks{akshay@cs.umass.edu}}
\author[1]{
Aarti Singh
\thanks{aarti@cs.cmu.edu}}

\affil[1]{Machine Learning Department\\
Carnegie Mellon University}
\affil[2]{Microsoft Research, New York City}

\maketitle

\begin{abstract}
This paper studies the problem of estimating the covariance of a collection of vectors using only highly compressed measurements of each vector.
An estimator based on back-projections of these compressive samples is proposed and analyzed.
A distribution-free analysis shows that by observing just a single linear measurement of each vector, one can consistently estimate the covariance matrix, in both infinity and spectral norm, and this same analysis leads to precise rates of convergence in both norms. 
Via information-theoretic techniques, lower bounds showing that this estimator is minimax-optimal for both infinity and spectral norm estimation problems are established.
These results are also specialized to give matching upper and lower bounds for estimating the population covariance of a collection of Gaussian vectors, again in the compressive measurement model.
The analysis conducted in this paper shows that the effective sample complexity for this problem is scaled by a factor of $m^2/d^2$ where $m$ is the compression dimension and $d$ is the ambient dimension.
Applications to subspace learning (Principal Components Analysis) and learning over distributed sensor networks are also discussed. 
\end{abstract}

\section{Introduction}
\label{sec:intro}

Covariance matrices provide second-order information between a collection of random variables and play a fundamental role in statistics and signal processing.
Concrete examples include dimensionality reduction, where covariance information is a sufficient statistic for the widely used Principal Components Analysis (PCA), and linear discriminant analysis, a popular classification method. 
An important statistical task is \emph{covariance estimation}, where the goal is to recover the population covariance matrix of a distribution, given independent and identically distributed samples.

In this paper, we study a variant of the covariance estimation problem, where the samples are observed only through low-dimensional random projections. 
This estimation problem has roots in compressed sensing, where random projections have been used to reduce measurement overhead associated with high-dimensional signals.
It is also motivated by problems in learning over distributed sensor networks, where both power and communication constraints may limit the measurement capabilities of a single sensor.
We describe this application in more detail in Section~\ref{sec:distributed}.

In the first part of the paper, we propose and analyze a covariance estimator based on these low-dimensional, compressed, observations. 
We specifically consider a model where an independent random projection is used for each data vector.
We show that even when each vector is observed only via projection onto a one-dimensional subspace (i.e., one linear measurement), one can consistently and accurately estimate the sample covariance matrix of the data vectors in both spectral and infinity norms. 
In our analysis, we make no distributional assumptions on the data vectors themselves and attempt to recover the sample covariance, since no population covariance exists.
We present a specialization of this distribution-free analysis to the case where the data vectors are drawn from a Gaussian distribution and the goal is to estimate the population covariance. 
As two motivating applications of our analysis, we give guarantees for subspace learning (or Principal Components Analysis) and for learning over distributed sensor networks.

In the second part of the paper, we consider the fundamental limits of this estimation problem. 
Using information-theoretic tools, we derive lower bounds for a variety of settings, including the distributional and distribution-free settings under which we analyze our estimator.
This analysis reveals that our covariance estimator achieves the minimax-rate for this problem up to constant and logarithmic factors, meaning that our estimator is essentially the best one can hope for.
We also consider an alternative popular measurement paradigm, where a single low-dimensional random projection is used for all data vectors and show that this approach is inconsistent for the covariance estimation problem. 

Our work deviates from the majority of work on compressive estimation in that we do not make structural assumptions on the estimand, in this case the target covariance. 
A number of papers assume that the target covariance is low rank~\cite{cai2015rop,chen2013exact}, sparse~\cite{dasarathy2013sketching}, or that the inverse covariance is sparse~\cite{kolar2012consistent,ravikumar2011high}.
The broad theme of this line of work is that when the target covariance has some low-dimensional structure, different sensing strategies and far fewer total measurements (via random projection) suffice to achieve the same error as direct observation in the unstructured case (See Section~\ref{sec:related} for details).
However, when the target covariance does not have low-dimensional structure, these methods can fail dramatically as we show with our lower bounds. 

In contrast, our work instead examines the statistical price for compression when the covariance matrix \emph{does not} exhibit any low-dimensional structure.
In the unstructured setting, compressing the samples requires that one use significantly more measurements to achieve comparable level of accuracy to the uncompressed case. 
We precisely quantify this increase in measurement, showing that the effective sample size shifts from $n$ to $nm^2/d^2$, where the projection dimension is $m$ and the ambient dimension is $d$. 
Since we must have $m \le d$, this means that one needs more samples to achieve a specified accuracy under our measurement model, in comparison with direct observation.
On the other hand, our results apply even when $m=1$, so consistent recovery is possible even when each sample is compressed down to a single scalar. 
This effective sample size is present in all of our upper and lower bounds, showing that indeed, there is a price to pay for compression without structural assumptions. 
Note that this quadratic growth in effective sample size also matches recent results on covariance estimation from missing data~\cite{kolar2012consistent,loh2012high}.

While our focus is on the unstructured case, we do show that our estimator can \emph{adapt} to structure present in the problem. 
Specifically, in the case where the data vectors lie on a $k$-dimensional subspace, the error bounds for our estimator match those of other approaches that specialize to this low rank setting~\cite{cai2015rop, chen2013exact}.
Thus, the simple estimator we introduce here addresses both structured and unstructured covariance estimation tasks.

Regarding proof techniques, our upper bounds are based on analysis of a carefully constructed \emph{unbiased} estimator for the target covariance matrix.
The natural estimator for this problem is asymptotically biased and hence inconsistent, but by exploiting properties of the Beta distributions that arises from random projections, we are able to analytically de-bias this natural estimator (See Section~\ref{sec:estimator}).
To obtain error bounds, we use concentration-of-measure arguments.
The challenge in this part of the analysis is that the relevant random variables have very large range even though their tails decay quite favorably; consequently, a straightforward application of a Bernstein-type inequality is too pessimistic.
In the $\ell_{\infty}$ case we avoid this issue with a conditioning argument, first showing that the random variables have much smaller range with high probability and then applying a Bernstein-type inequality conditioned on this event.
In the spectral norm case, we use a more powerful deviation bound (The Subexponential Matrix Bernstein inequality~\cite{tropp2011user}) that exploits sharper decay on all moments of the relevant random variables. 

For the lower bounds, our main technical contribution is a \emph{strong data-processing inequality}~\cite{anantharam2013maximal,duchi2013local,raginsky2014strong} which upper bounds the Kullback-Leibler divergence between two compressed Gaussian distributions by a small (less than one) multiple of the KL-divergence before compression.
This contraction in KL-divergence, in concert with a standard approach for establishing minimax lower bounds known as Fano's method, gives the lower bounds in this paper.

The remainder of this paper is organized as follows: We conclude this section with a formal specification of the covariance estimation problem and the observation model. 
In Section~\ref{sec:related}, we mention related results on covariance estimation and matrix approximation.
In Section~\ref{sec:estimator}, we develop our covariance estimator, providing a theoretical analysis in Section~\ref{sec:upper}.
Section~\ref{sec:upper} also contains some simulations results and a discussion of applications to subspace learning and learning in distributed sensor networks. 
We present all of our lower bounds in Section~\ref{sec:lower}.
All proofs of our theorems are in Section~\ref{sec:proofs} with a brief discussion in Section~\ref{sec:conclusion}.
Several technical lemmas are deferred to the appendices.

\subsection{Setup}
Let $x_1, \ldots, x_n$ be a collection of vectors in $\RR^d$ and define the covariance $\Sigma \triangleq \frac{1}{n}\sum_{t=1}^n x_t x_t^T$. 
We make no distributional assumptions on the sequence $\{x_t\}_{t=1}^n$ and therefore aim to recover the \emph{sample covariance} $\Sigma$, since there is no well-defined population version.
In particular, the sequence could be adversarially generated. 
When we specialize to the distributional setting, we will assume that the sequence $x_1, \ldots, x_n \sim \Ncal(0, \Sigma)$, where $\Sigma$ is the population covariance. 
Whether $\Sigma$ refers to the sample covariance in the distribution-free setting or the population covariance in the distributional setting will be clear from context. 
Let $X \in \RR^{d\times n}$ be a matrix whose $t^{\textrm{th}}$ column is the data vector $x_t$. 

Independently for all $t$, let $A_t \in \RR^{d\times m}$ be an orthonormal basis for an $m$-dimensional subspace drawn uniformly at random. 
We are interested in estimating $\Sigma$ from the observations $\{(A_t, A_t^Tx_t)\}_{t=1}^n$, so that each vector is compressed from $d$ dimensions down to $m$ dimensions.
Note that this measurement scheme is equivalent, in an information-theoretic sense, to drawing $m$-dimensional orthogonal projections $\Phi_t \in \RR^{d\times d}$ uniformly at random, and independently for all $t$, and observing $\{(\Phi_t, \Phi_t x_t)\}_{t=1}^n$. 
This equivalence can be easily seen by noting that the matrix $A_tA_t^T$ is a uniform-at-random $m$-dimensional orthogonal projection, while a uniform-at-random orthonormal basis for the subspace encoded by $\Phi_t$ has the same distribution as $A_t$. 
In both cases the vectors $x_t$ have been compressed down to $m$ dimensions. 

As terminology, we use the phrases ``data sequence" and ``samples" to denote the vectors $x_1, \ldots, x_n$, which we emphasize are only observed via compression.
We use ``observations" for the equivalent representations $(A_t, A_t^Tx_t)$ and $(\Phi_t, \Phi_tx_t)$.
We reserve the word ``measurements" for the linear operators $A_t$ or $\Phi_t$, which act on the data sequences to produce the observations. 
The term ``sample complexity" refers to the number of observations $n$ as a function of the parameters $m$ and $d$ required to achieve a desired error for a particular task\footnote{Technically, sample complexity refers to the number of observations $n$ as a function of $m,d,\epsilon$, and $\delta$ that suffice to achieve error $\epsilon$ with probability at least $1-\delta$ for a particular estimation task, but we often use this phrase loosely and suppress dependence on $\epsilon$ and $\delta$.}.

For a matrix $M$, let $\|M\|_F \triangleq \sqrt{\sum_{i,j=1}^d M_{i,j}^2}$ denote the Frobenius norm and let $\|M\|_{\infty} \triangleq \max_{i,j} |M_{i,j}|$ denote the element-wise infinity norm. 
We also use $\|M\|_{p,q}$ to denote the $\ell_{p,q}$ mixed norm, which is $\ell_q$ norm of the $\ell_p$ norms of the columns of the matrix.
For example, $\|M\|_{2,\infty} \triangleq \max_{j} \|m_j\|_2$ for a matrix with columns $m_j$, and this specific norm appears several times in our analysis.
For a symmetric matrix $M \in \RR^{d\times d}$ let $\|M\|_2 \triangleq \max_{x: \|x\|_2 = 1} |x^TMx|$ denote the spectral or operator norm.
For symmetric matrices $A,B$, we write $A \preceq B$ if $B-A$ is positive semidefinite, that is $x^T(B-A)x \geq 0$ for all $x$. 
For a vector $v$, $\|v\|_2$, $\|v\|_{\infty}$ are the Euclidean and infinity norm, respectively. 
In $d$-dimensions, we use $e_1,\ldots,e_d$ to denote the standard basis elements. 
Lastly, we use the standard Big-O and Little-O notation for asymptotic characterizations, where $\tilde{O}$ and $\tilde{\Omega}$ suppress dependence on logarithmic factors.

\section{Related Work}
\label{sec:related}

Estimating a covariance matrix from samples is a classical problem in statistics with applications across the spectrum of scientific disciplines.
Recent work has focused on the high-dimensional setting, where the dimensionality of the data points is large relative to the number of samples.
In this setting, a number of structural assumptions that lead to tractable estimators have been proposed and studied. 

One approach to these estimation problems is through \emph{compressive sensing}~\cite{donoho2006compressed,candes2006robust}, where the data is observed through low-dimensional random projections from which the estimand can be algorithmically recovered.
These ideas have been extended to matrices in a line of work that studies matrix recovery and covariance estimation from compressive measurements~\cite{cai2015rop, chen2013exact, dasarathy2013sketching}.
The problem we study deviates from these results in two important ways: (1) we make no structural assumptions about the underlying covariance matrix where prior work assumes low-rank structure or sparsity, and (2) we obtain compressive measurements of the individual samples $x_t$ rather than the covariance matrix directly. 

Directly measuring the covariance matrix can be implemented in our setting as a shared compression operator $A \in \RR^{d\times m}$ with the observations $\{A^Tx_t\}_{t=1}^n$. 
Unfortunately, in the absence of structure, this approach does not allow for non-trivial compression as it requires $m = \Theta(d)$, which is demonstrated by the upper bounds in prior work~\cite{cai2015rop,chen2013exact,dasarathy2013sketching} and our minimax lower bounds (see Proposition~\ref{prop:fixed}).
On the other hand, our approach of independently compressing each sample $x_t$ is a more flexible measurement scheme that cannot be implemented in the matrix recovery setting, but it does enable non-trivial covariance estimation even in the absence of structure. 
Moreover, in the presence of structural assumptions, our estimator does achieve statistical error that is comparable with more specialized approaches (See Corollary~\ref{cor:low_rank_adapt}).

Cai and Zhang~\cite{cai2015rop} also study a setting where data vectors $x_t$ are drawn from a Gaussian distribution with spiked covariance ($\Sigma = I + M$ where $M$ is low rank) and compressed independently, so only $(a_t, a_t^Tx_t)$ is observed for uniform at random unit vectors $\{a_t\}_{t=1}^n$.
While they do make a spiked covariance assumption, their Theorem 4.1 does lead to a meaningful guarantee even when $M$ is full rank, in which case it can be viewed as complementary to our results.
The main differences are that we derive spectral and infinity norm bounds while they obtain a Frobenius norm bound and that we consider more general compression dimension while they focus on $m=1$. 
Additionally, our estimator is conceptually simpler as it does not rely convex programming.

The only other results that do not require structural assumptions are those of Pourkamali-Anarki and Hughes~\cite{pourkamali2014memory} and our previous work~\cite{krishnamurthy2014subspace}. 
As in the present work, Pourkamali-Anarki and Hughes~\cite{pourkamali2014memory} study the covariance estimation problem when samples are observed only through low-dimensional linear measurements. 
Their main interest is in using sparse measurement matrices to study a computational-statistical tradeoff for this problem, but in the dense case most similar to ours, their estimator can be used for consistent estimation of the principal components.
However, they do not characterize the rate of convergence and it is inconsistent for the target covariance.
In contrast, our work studies a consistent estimator for the target covariance, and our main interest is in precisely characterizing the convergence rate for the covariance estimation task.
Additionally, our work makes much weaker assumptions on the data (specifically they make distributional assumptions while we operate in an adversarial setting), while their work considers a broader class of linear measurements, although their class does not contain the random projections we study.

Our previous work~\cite{krishnamurthy2014subspace} focuses on the compressive subspace learning problem but does provide guarantees for covariance estimation.
There are, however, two main shortcomings that we resolve in the present paper.
First, the estimator there is based on a version of data splitting, and, consequently, it does not allow for the $m=1$ case as we do here.
Secondly, the rates of convergence are worse in our previous work, whereas we show that the rates derived in this paper are minimax optimal.

In the theoretical computer science and numerical linear algebra literature, there are several works that use random projections for the purposes of fast approximation to the singular value decomposition of an unstructured matrix~\cite{sarlos2006improved,halko2011finding,liberty2007randomized}.
In this matrix approximation problem, we fully observe a data matrix $X \in \RR^{d \times n}$, and the goal is to efficiently compute a rank $k$ matrix $\hat{X}$ with the guarantee that $\|X - \hat{X}\| \le f(\|X - X_k\|)$ for some norm (usually spectral or Frobenius) and some function $f$, where $X_k$ is the best rank-$k$ approximation to $X$.
The main differences with our setting are (a) that the data matrix is fully observed, (b) the low-rank versus unstructured nature, and (c) that the performance measure is on the data matrix $X$ rather than the covariance $\Sigma$.
Consequently, state-of-the-art matrix approximation algorithms provably fail at covariance estimation (as we show in Proposition~\ref{prop:fixed}), while the procedure we develop for covariance estimation does not achieve state-of-art matrix approximation performance.

Despite these differences, it is worth briefly discussing algorithmic ideas in the matrix approximation literature.
For concreteness, consider the algorithm of Halko, Martinsson, and Tropp~\cite{halko2011finding}, which is a representative example from this line of research.
Their algorithm first right-multiplies $X$ by a small random matrix $R$ to obtain a matrix $Y = XR$ that approximates the column space of $X$, and then it projects the columns of $X$ onto the subspace spanned by $Y$ to obtain the estimator $\hat{X} = YY^TX$.
While this scheme leads to high-quality matrix approximations, it both pre- and post-multiplies the matrix $X$, which amounts to obtaining compressive measurements of both the rows and columns of the data and is not possible in our setting..
Therefore these algorithms do not address our problem.

Another closely related line of work focuses on matrix recovery from missing data. 
While the majority of the results here focus on low rank matrices or other structured settings~\cite{recht2011simpler,negahban2012restricted}, there have been recent results focusing explicitly on the covariance estimation problem in the unstructured setting.
For example, Kolar and Xing~\cite{kolar2012consistent} consider a setting where each coordinate of each data vector is missing with probability $1-\alpha$, independently from other coordinates and vectors. 
They show that to estimate the covariance matrix in $\ell_{\infty}$ norm, the effective sample size shrinks from $n$ to $n\alpha^2$, and they propose an estimator that can be used to learn the structure of a Gaussian graphical model.
Gonen \emph{et al.}~\cite{gonen2014sample} study the subspace learning problem in the missing data setting, while Loh and Wainwright~\cite{loh2012high} consider high-dimensional linear regression with missing data.
Both works show that the sample complexity is increased by a factor related to the squared fraction of entries observed per column.
While the measurement model in our work is different, qualitatively our results are similar to these; since $m/d = \alpha$, we show a similar increase in sample complexity when the data is observed via random projection. 

While the statistical behavior is similar in these two settings, obtaining sharp error bounds for the compressed case is analytically much more challenging than the missing data one. 
In the missing data case, it is common (in fact, necessary) to assume that the data vectors are incoherent or have small $\ell_{\infty}$ norm relative to their $\ell_2$ norm (The typical assumption is $d\|x_t\|_{\infty}^2 \le \mu \|x_t\|_2^2$ for some constant $\mu$~\cite{recht2011simpler}). 
One can obtain error bounds for uniform-at-random coordinate sampling under this assumption simply by application of a Bernstein-type inequality, as all of the relevant random variables have range and variance on the order of $\frac{m}{d}\|x_t\|_{2}^2$ if $m$ measurements per sample are obtained. 
On the other hand, the measurement process in the compressive model leads to random variables with range as large as $\|x_t\|_{2}^2$, so that a na\"{i}ve application of these concentration inequalities yields very weak error bounds.
Our analysis therefore uses two more sophisticated approaches: for the $\ell_{\infty}$-norm bound we use a two-stage conditioning argument, and for the spectral norm bound we use a more refined deviation bound that exploits the sharper tail decay of our random variables. 

\section{The Covariance Estimator}
\label{sec:estimator}

In this section, we develop our compressive covariance estimator.
The estimator is based on an adjustment to the observed covariance, i.e. the covariance of the $\Phi_t x_t$ vectors.
The adjustment is motivated by a characterization of the bias of the observed covariance. 

Specifically, let $\hat{\Sigma}_1 \triangleq \frac{d^2}{nm^2} \sum_{t=1}^n (\Phi_tx_t)(\Phi_t x_t)^T$ be a rescaled version of the observed covariance. 
$\hat{\Sigma}_1$ is an intuitive estimator for $\Sigma$, but, as we will see, it is biased even as $n$ tends to infinity, meaning that $\hat{\Sigma}_1$ is \emph{not} a consistent estimator for $\Sigma$.
Instead, our estimate for the sample covariance $\Sigma$ is
\begin{align}
\label{eq:estimator}
\hat{\Sigma} \triangleq \frac{m \left((d+2)(d-1) \hat{\Sigma}_1 - (d-m)\tr(\hat{\Sigma}_1)I_d\right)}{d(dm+d-2)}.
\end{align}

This estimator is a de-biased version of $\hat{\Sigma}_1$. 
Unbiasedness is helpful for our analysis, but also in applications, since our deviation bounds immediately lead to valid confidence intervals.
For a biased estimator, the confidence interval must incorporate an upper bound on the bias, so it may be considerably wider in practice.

The specification of $\hat{\Sigma}$ is motivated by the following proposition, which analytically characterizes the bias of $\hat{\Sigma}_1$ based on properties of the Beta distribution (See Fact~\ref{fact:beta}). The proof is deferred to Section~\ref{sec:proof_unbiased}.
\begin{proposition}[De-biasing]
\label{prop:unbiasedness}
Let $\Sigma = \frac{1}{n}\sum_{t=1}^n x_t x_t^T$ and $\hat{\Sigma}_1 = \frac{d^2}{m^2}\frac{1}{n}\sum_{t=1}^n (\Phi_tx_t)(\Phi_t x_t)^T$.
Then
\ifthenelse{\equal{\numcolumns}{one}}{
\begin{align*}
\EE \hat{\Sigma}_1 = \frac{d(dm+d-2)\Sigma + d(d-m)\tr(\Sigma)I_d}{m(d+2)(d-1)}. 
\end{align*}
}{
\begin{align*}
\EE \hat{\Sigma}_1 = \frac{d(dm+d-2)\Sigma + d(d-m)\tr(\Sigma)I_d}{m(d+2)(d-1)}. 
\end{align*}
}
\end{proposition}

With this expansion of the bias, it is also easy to see that $\tr(\EE\hat{\Sigma}_1) = \frac{d}{m} \tr(\Sigma)$.
Substituting in for $\tr(\Sigma)$ and re-arranging, we see that
\begin{align}
\label{eq:unbiased_check}
\Sigma = \frac{m\left( (d+2)(d-1) \EE\hat{\Sigma}_1 - (d-m)\tr(\EE\hat{\Sigma}_1)I_d\right)}{d(dm+d-2)}.
\end{align}
Since trace is a linear operator, we immediately see that our estimator is unbiased for $\Sigma$. 

\section{Upper Bounds and Consequences}
\label{sec:upper}

We now turn to our analysis of the estimator $\hat{\Sigma}$. 
In this section, we upper bound the error in both spectral and $\ell_{\infty}$ norms for $\hat{\Sigma}$ in the distribution-free setting.
We also specialize these results to the problem of estimating the population covariance of a sequence of independent and identically distributed Gaussian vectors. 
Lastly, we present a brief numerical simulation and discuss applications to subspace learning and learning in distributed sensor networks.

Our first two theorems give error bounds for our estimator, in entry-wise $\ell_{\infty}$ and spectral norm, respectively. 
\begin{theorem}[$\ell_{\infty}$ Upper Bound]
\label{thm:infinity_upper}
Let $d \ge 2$ and $\delta \in (0,1)$ such that $\delta \ge 4d^2 \exp\left(-n/9600\right)$. 
There exist a universal constant $\kappa > 0$ such that, with probability at least $1-\delta$,
\ifthenelse{\equal{\numcolumns}{one}}{
\begin{align*}
\|\hat{\Sigma} - \Sigma\|_{\infty} &\le
 \kappa\|X\|_{\infty}^2\left(\sqrt{\frac{d^2\log^3\left(\frac{nd}{\delta}\right)}{nm^2}}
 + \frac{d^2\log^2\left(\frac{nd}{\delta}\right)}{nm^2}\right).
\end{align*}
}{
\begin{align*}
\|\hat{\Sigma} - \Sigma\|_{\infty} &\le
 \kappa\|X\|_{\infty}^2\left(\sqrt{\frac{d^2\log^3\left(\frac{nd}{\delta}\right)}{nm^2}}
 + \frac{d^2\log^2\left(\frac{nd}{\delta}\right)}{nm^2}\right).
\end{align*}
}
\end{theorem}

\begin{theorem}[Spectral Upper Bound]
\label{thm:spectral_upper_tight}
Let $d \ge 2$ and define
\begin{align*}
S_1 \triangleq \left\|\frac{1}{n}\sum_{t=1}^n \|x_t\|_2^2x_tx_t^T\right\|_2, \quad \textrm{and} \quad S_2 \triangleq \frac{1}{n}\sum_{t=1}^n \|x_t\|_2^4.
\end{align*}
There exists universal constants $\kappa_1, \kappa_2 > 0$ such that for any $\delta \in (0,1)$, with probability at least $1-\delta$,
\begin{align*}
\ifthenelse{\equal{\numcolumns}{one}}{
\|\hat{\Sigma} - \Sigma\|_2 &\le \kappa_1 \left(\sqrt{\frac{d}{m}S_1} +\sqrt{\frac{d}{m^2}S_2}\right)\sqrt{\frac{\log(d/\delta)}{n}} + \kappa_2\frac{d\|X\|_{2,\infty}^2}{nm}\log(d/\delta).
}{
\|\hat{\Sigma} - \Sigma\|_2 &\le \kappa_1 \left(\sqrt{\frac{d}{m}S_1} +\sqrt{\frac{d}{m^2}S_2}\right)\sqrt{\frac{\log(d/\delta)}{n}}\\
& + \kappa_2\frac{d\|X\|_{2,\infty}^2}{nm}\log(d/\delta).
}
\end{align*}
\end{theorem}

We defer the proofs of both theorems to Section~\ref{sec:proofs}.
Note that both theorems are distribution-free; we make no assumptions on the data.
In particular, the data sequence can be adversarially generated or come from a heavy-tailed distribution, but observe that some sequence-dependent quantities do appear in the deviation bounds. 
In both theorems, the assumption on $d$ is very mild, as the $d=1$ case reduces to uncompressed covariance estimation, since $m$ must be at least $1$. 
In Theorem~\ref{thm:infinity_upper}, the assumption on $\delta$ is also mild as the lower bound is decaying exponentially with $n$.

To interpret Theorem~\ref{thm:spectral_upper_tight}, notice that $S_1 \le \|X\|_{2,\infty}^2\|\Sigma\|_2$ and $S_2 \le d \|X\|_{2,\infty}^2\|\Sigma\|_2$. 
With both of these bounds and when $n$ is sufficiently large, the error guaranteed by the theorem has a leading term of order $\tilde{O}\left(\sqrt{\frac{d^2\|X\|_{2,\infty}^2\|\Sigma\|_2}{nm^2}}\right)$.
We leave the dependence on $S_1$ and $S_2$ explicit in the statement because much sharper bounds on these two quantities are often possible. 
In particular, we will show that when the target covariance is low rank, one can obtain a refined spectral norm bound.

To compare with existing work, it is best to specialize to the case where the data vectors come from a Gaussian distribution. 
Application to other data distributions, including heavy-tailed distributions, is straightforward with tail bounds on $\|X\|_{2,\infty}$ and $\|X\|_{\infty}$, but out of scope for this paper.
Standard tail bounds on $\|X\|_{2,\infty}$ and $\|X\|_{\infty}$ in the Gaussian case yield the following:

\begin{corollary}[Gaussian Upper Bounds]
\label{cor:gaussian_upper}
Let $x_1, \ldots, x_n \sim \Ncal(0, \Sigma)$ and construct $\hat{\Sigma}$ as in Equation~\eqref{eq:estimator}. 
Then for any $\delta \in (0,1)$, there exist universal constants $\kappa_1, \kappa_2 > 0$ such that, with probability at least $1 - \delta$,
\ifthenelse{\equal{\numcolumns}{one}}{
\begin{align*}
\|\hat{\Sigma} - \Sigma\|_{\infty} & \le \kappa_1 \|\Sigma\|_{\infty} \left(\sqrt{\frac{d^2\log^5(nd/\delta)}{nm^2}} + \sqrt{\frac{\log(d/\delta)}{n}} + \frac{d^2\log^3(nd/\delta)}{nm^2}\right),
\\
\|\hat{\Sigma} - \Sigma\|_2 & \le \kappa_2 \|\Sigma\|_2 \left( \sqrt{\frac{d^3\log^2(nd/\delta)}{nm^2}} + \frac{d^3\log^2(nd/\delta)}{nm^2} + \sqrt{\frac{d\log(1/\delta)}{n}}\right).
\end{align*}
}{
\begin{align*}
\|\hat{\Sigma} - \Sigma\|_{\infty} & \le \kappa_1 \|\Sigma\|_{\infty} \left(\sqrt{\frac{d^2\log^5(nd/\delta)}{nm^2}} + \sqrt{\frac{\log(d/\delta)}{n}}\right.\\
& \left. \qquad + \frac{d^2\log^3(nd/\delta)}{nm^2}\right),
\\
\|\hat{\Sigma} - \Sigma\|_2 & \le \kappa_2 \|\Sigma\|_2 \left( \sqrt{\frac{d^3\log^2(nd/\delta)}{nm^2}} + \frac{d^3\log^2(nd/\delta)}{nm^2}\right.\\
&\left. \qquad + \sqrt{\frac{d\log(1/\delta)}{n}}\right).
\end{align*}
}
The first bound holds when $d \ge 2$ and $\delta \ge 4d^2 \exp(-n/9600)$, while the second bound holds when $d\ge 2$ and $n \ge d\log(1/\delta)$.
\end{corollary}

Here we make several remarks:
\begin{enumerate}
\item The requirement $n \ge d \log(1/\delta)$ is not necessary but leads to a simpler bound.
  However, when $n < d\log(1/\delta)$ it is impossible to achieve non-trivial spectral norm error even in the uncompressed case, so the requirement is not particularly strong. 
\item When $n$ is large relative to $d^2/m^2$ and ignoring logarithmic factors, the leading terms in the error bounds are $\tilde{O}\left(\|\Sigma\|_{\infty}\sqrt{\frac{d^2}{nm^2}}\right)$ in $\ell_{\infty}$ norm and $\tilde{O}\left(\|\Sigma\|_2\sqrt{\frac{d^3}{nm^2}}\right)$ in spectral norm. 
  In comparison, to estimate the population covariance of a Gaussian distribution when the vectors $\{x_t\}_{t=1}^n$ are directly observed, it is well known that the sample covariance achieves rates $\tilde{O}\left(\|\Sigma\|_{\infty}\sqrt{\frac{1}{n}}\right)$ and $\tilde{O}\left(\|\Sigma\|_2\sqrt{\frac{d}{n}}\right)$ in infinity and spectral norm respectively~\cite{vershynin2010introduction}. 
  Thus, the effective sample size shrinks from $n$ to $nm^2/d^2$ in the compressed setting. 
\item Apart from our previous work~\cite{krishnamurthy2014subspace}, this is the first estimator with such strong guarantees in the compressed setting.
As we mentioned, most existing work focuses on recovery under strong structural assumptions of the population covariance, for example low rank~\cite{chen2013exact,cai2015rop} or spiked covariance~\cite{cai2015rop}.
\item Nevertheless, we can show that our estimator \emph{adapts} to low dimensional structures exhibited by the target covariance.
In the Gaussian case, if the target covariance has rank at most $k$, then an estimator $\hat{\Sigma}_k$ formed by zero-ing out all but the largest $k$ eigenvalues of $\hat{\Sigma}$ has a much more favorable convergence rate:
\begin{corollary}
\label{cor:low_rank_adapt}
Consider the same setting as Corollary~\ref{cor:gaussian_upper} with $d \ge 2$, but further assume that $\textrm{rank}(\Sigma) \le k$.
Then, there exists a universal constant $\kappa > 0$ such that for any $\delta \in (0,1)$, when $n \ge d \log(1/\delta)$, with probability $1 - \delta$, we have
\begin{align*}
\ifthenelse{\equal{\numcolumns}{one}}{
\|\hat{\Sigma}_k - \Sigma\|_2 \le \kappa \|\Sigma\|_2 \left(\sqrt{\frac{dk^2\log^2(nd/\delta)}{nm^2}} + \sqrt{\frac{k\log(1/\delta)}{n}} + \frac{dk\log^2(nd/\delta)}{nm}\right).
}{
\|\hat{\Sigma}_k - \Sigma\|_2 &\le \kappa \|\Sigma\|_2 \left(\sqrt{\frac{dk^2\log^2(nd/\delta)}{nm^2}}\right.\\
 &\left.+ \sqrt{\frac{k\log(1/\delta)}{n}}+ \frac{dk\log^2(nd/\delta)}{nm}\right).
}
\end{align*}
\end{corollary}

This is the low-rank version of the spectral norm bound in Corollary~\ref{cor:gaussian_upper}; the only additional assumption is that the target covariance has rank at most $k$. 
The bound shows that if $n = \tilde{\Theta}(\max\{d, k/\epsilon^2\})$, we may set $m = O(k\log^2(d)/\epsilon)$ to achieve spectral norm error $O(\epsilon)$, which agrees with the sample complexity bounds in recent results~\cite{chen2013exact,cai2015rop,recht2011simpler}.

\item In comparison with our previous work~\cite{krishnamurthy2014subspace}, the results here are significantly more refined. 
First, our previous work used a data-splitting technique to avoid the bias demonstrated in Proposition~\ref{prop:unbiasedness} and consequently did not address the $m=1$ case as we do here.
Secondly, in terms of rates, the results here are sharper. 
In Krishnamurthy \emph{et al.}~\cite{krishnamurthy2014subspace}, the rate of convergence in spectral norm for the full rank Gaussian case has leading order term $\tilde{O}\left(\|\Sigma\|_2\sqrt{\frac{d^3}{nm}}\right)$ which is polynomially worse than our $\tilde{O}\left(\|\Sigma\|_2\sqrt{\frac{d^3}{nm^2}}\right)$ bound\footnote{That paper has an error that changes their reported rate to $\tilde{O}\left(\sqrt{\frac{d \mu^2}{nm}} + \frac{d \mu}{nm^2}\right)$ where $\mu$ upper bounds the squared Euclidean norm of each vector. In the Gaussian case, $\mu = \tilde{O}(d\|\Sigma\|_2)$.}.
As we will see in Theorem~\ref{thm:spectral_lower} below, this latter rate is minimax optimal. 
Finally, we also derive $\ell_{\infty}$ guarantees, whereas our prior work focuses on the spectral norm. 
\end{enumerate}

\begin{figure*}
\begin{center}
\includegraphics[scale=\figsizeb]{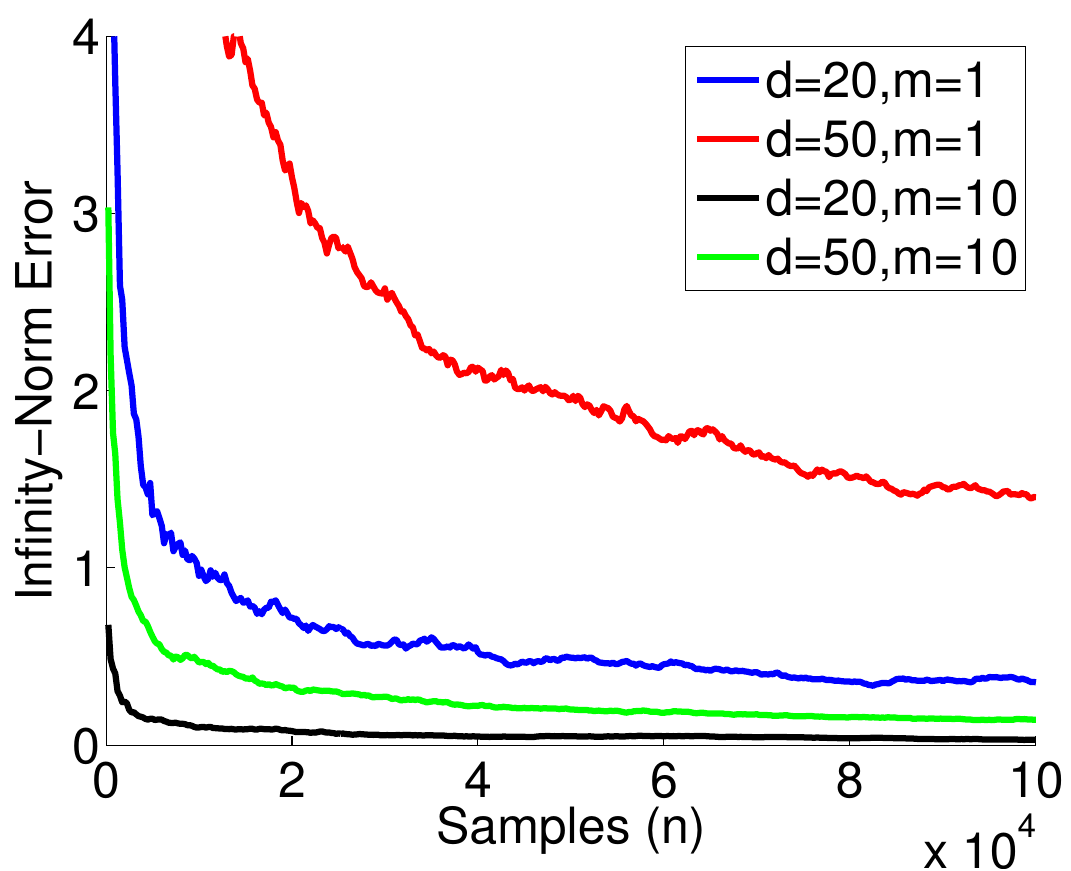}
\includegraphics[scale=\figsizeb]{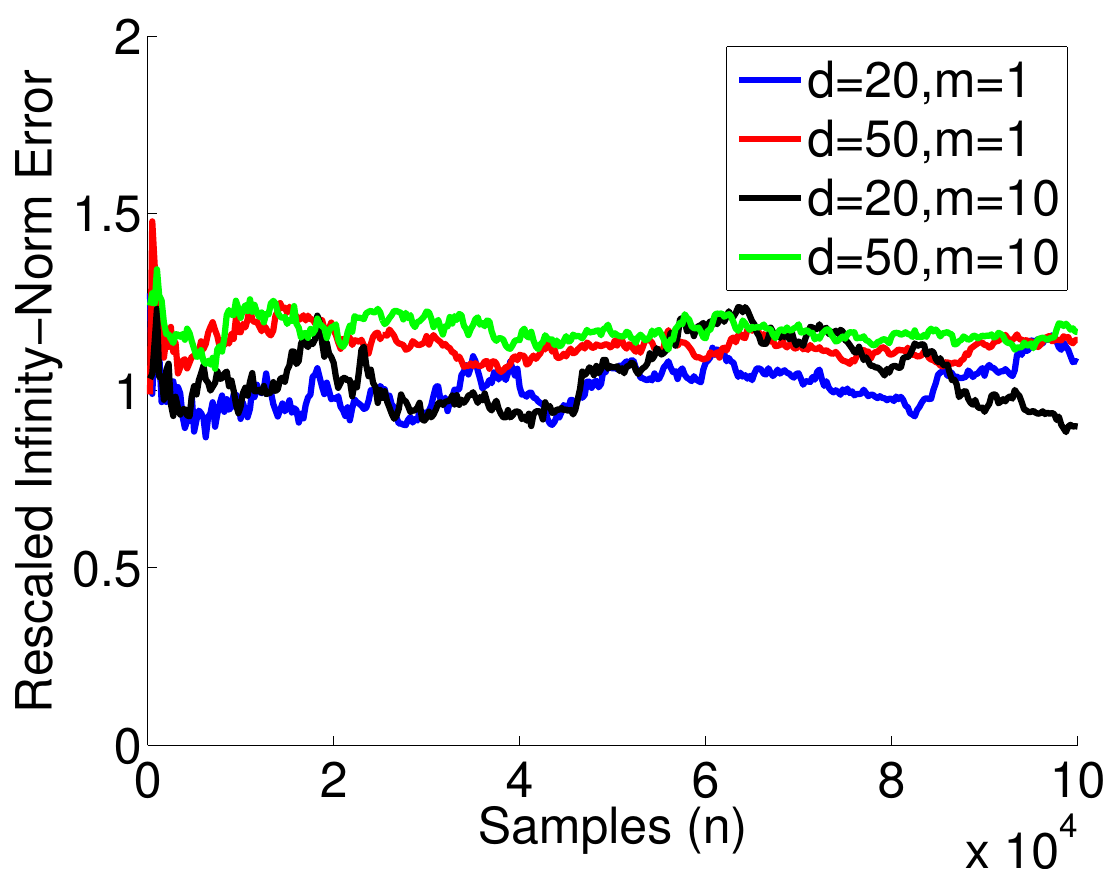}
\includegraphics[scale=\figsizeb]{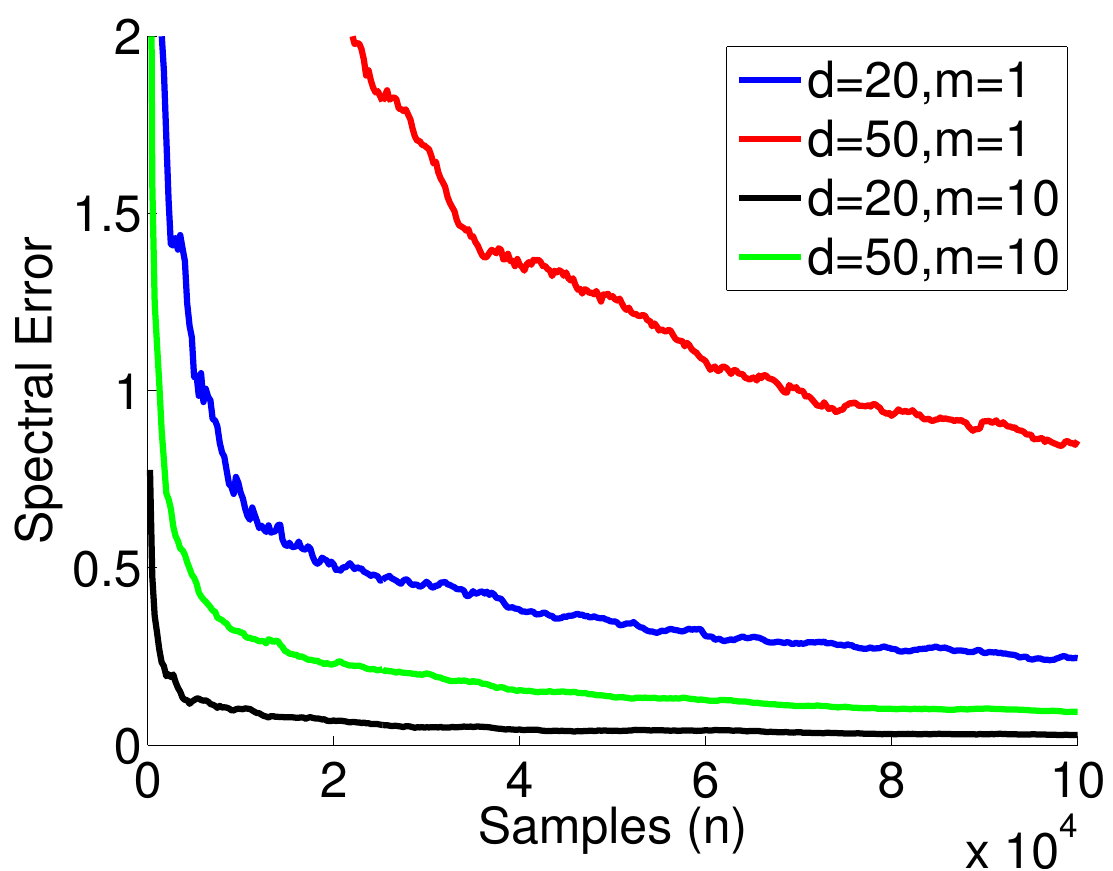}
\includegraphics[scale=\figsizeb]{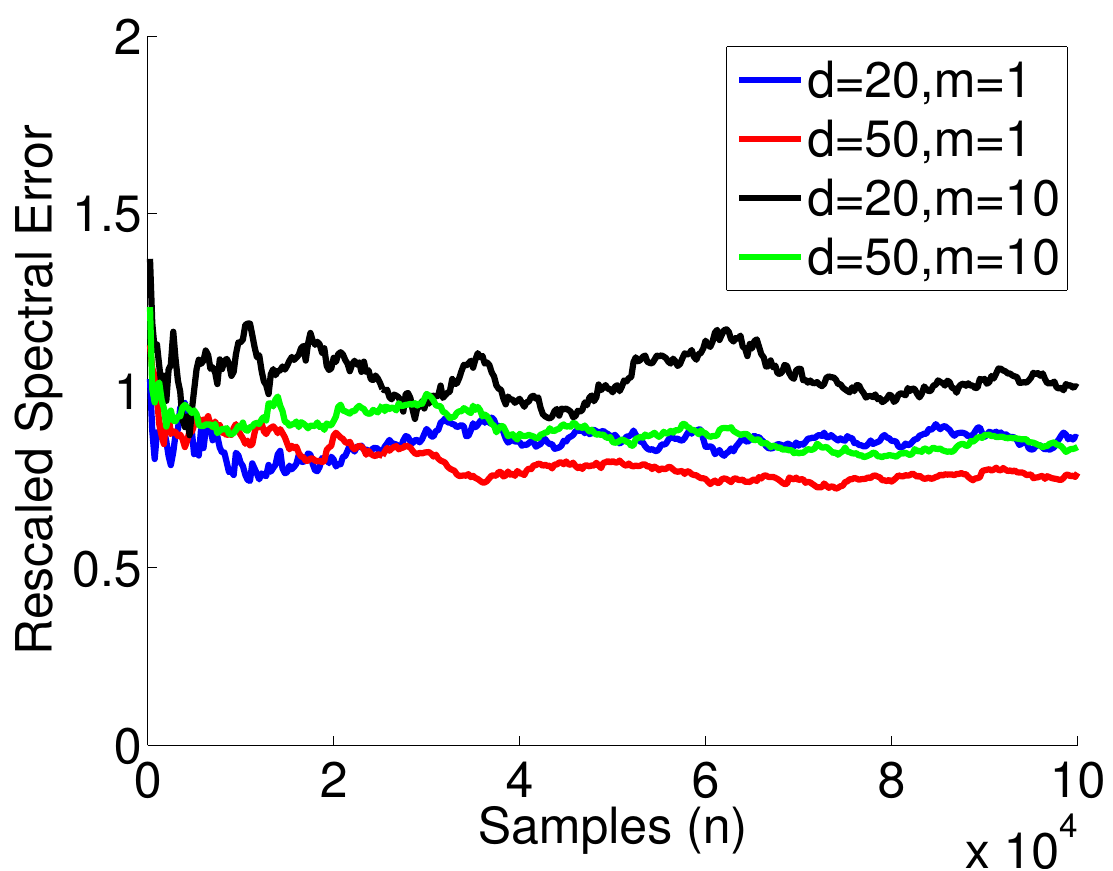}
\end{center}
\caption{Rates of convergence for our compressed covariance estimator alongside rescaled rates for infinity and spectral norm and different settings of $d,m$. 
Rescaling factor is $\sqrt{\frac{n m^2}{d^2\log^3(d)}}$ and $\sqrt{\frac{nm^2}{d^3}}$ for infinity norm (left two panels) and spectral norm (right two panels), respectively.}
\label{fig:rate_exp}
\end{figure*}

The proof of these results are based on showing that $\hat{\Sigma}_1$ concentrates sharply around its mean.
For both results, a crude application of exponential deviation bounds does not suffice, as the relevant random variables have large range, although the tails decay quite favorably.
We therefore use a more refined analysis to exploit this sharp tail decay. 
For the infinity norm bound, we use a conditioning argument where we first provide a probabilistic bound on the range of the random variables and then apply the Bernstein inequality conditioned on this event.
For the spectral norm bound, we instead use properties of Beta random variables to upper bound all of the moments in terms of the quantities $S_1$ and $S_2$ and then apply the Subexponential Matrix Bernstein inequality~\cite{tropp2011user}.
The corollaries are based on using well-known Gaussian concentration inequalities to bound the sample-dependent quantities in the theorems.

\subsection{Synthetic Experiments}
In Figure~\ref{fig:rate_exp}, we empirically validate the error bounds in Corollary~\ref{cor:gaussian_upper} by recording the infinity and spectral norm error of our estimator across several problem parameters. 
In the left two panels, the data is drawn from a multivariate Gaussian distribution whose covariance matrix has unit infinity norm, while in the right two panels, the data is normally distributed and the covariance matrix has unit spectral norm.
We plot the infinity norm error (left most) as a function of the number of samples alongside the rescaled infinity norm error $\|\hat{\Sigma} - \Sigma\|_\infty \sqrt{\frac{nm^2}{d^2\log^3d}}$ (second from left).
Similarly, in the right two panels, we plot the spectral norm error (second from right) alongside the rescaled error $\|\hat{\Sigma} - \Sigma\|_2 \sqrt{\frac{nm^2}{d^3}}$ (right most).

The experiment confirms Corollary~\ref{cor:gaussian_upper}, namely that our covariance estimator enjoys an error bound that converges to zero with $n$ in both infinity and spectral norm. 
In addition, the curves in the second and fourth plot validate the error bounds in two ways.
First, the fact that the curves in the second and fourth plots are flat validates that we have accurately captured the dependence between the error and the number of samples $n$, confirming the $n^{-1/2}$ convergence rate.
Secondly, the fact that these curves are tightly clustered suggests that we have also captured the dependence on $d$ and $m$ so that, modulo logarithmic factors, our bounds are sharp for this estimator. 

We also compare our approach with an algorithm that uses the same random projection for each vector.
There are many algorithms of this form~\cite{halko2011finding,cai2015rop,chen2013exact,dasarathy2013sketching}, and, as a representative example, we use the algorithm of Halko, Martinsson, and Tropp~\cite{halko2011finding}, which we described in Section~\ref{sec:related}.
Recall that this algorithm operates on both the rows and the columns of the matrix, so cannot actually be deployed in applications where the the data vectors (columns) are observed in a compressed fashion.
We emphasize that this algorithm is designed for low-rank matrix approximation rather than covariance estimation and that the objectives in these tasks are considerably different.
This comparison is more a demonstration that a shared compression operator is not suitable for unstructured covariance estimation than a criticism of their algorithm, or of matrix approximation more broadly.

\begin{figure}
\begin{center}
\includegraphics[scale=\figsizea]{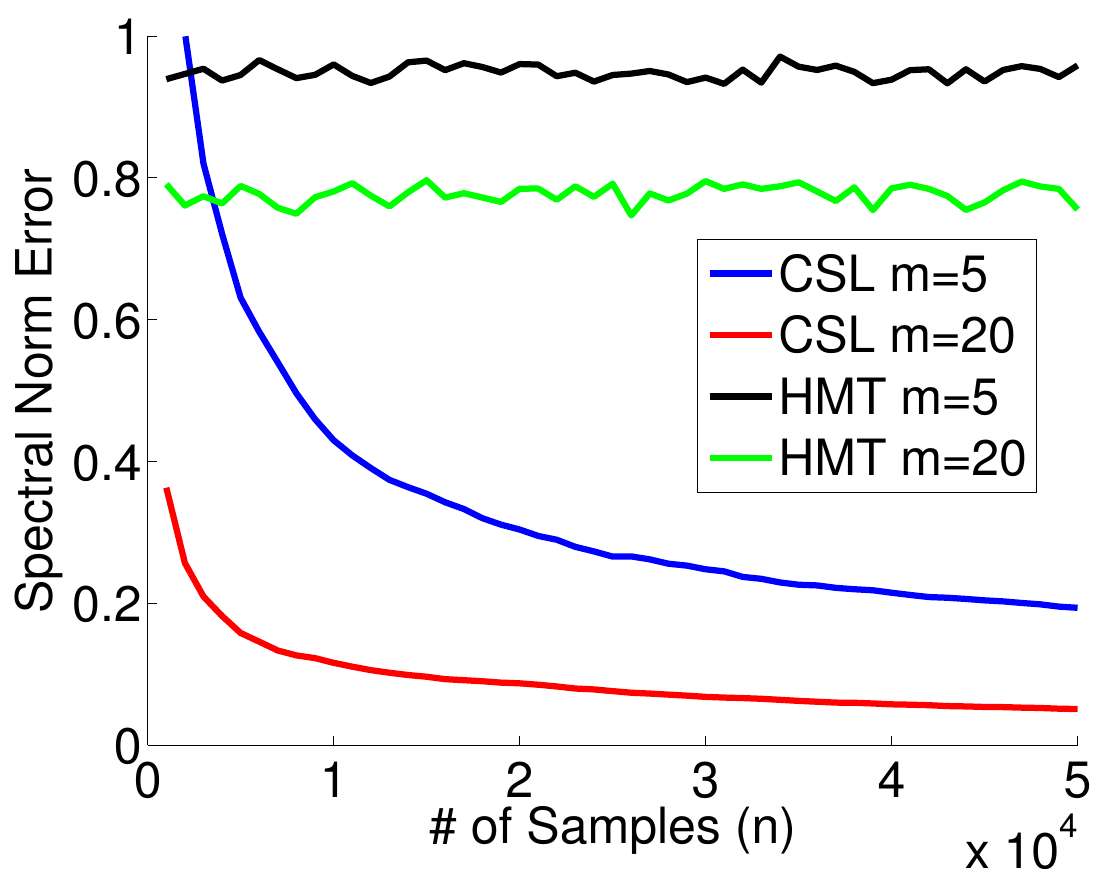}
\end{center}
\caption{Comparison of our approach (CSL) against the algorithm of Halko, Martinsson, and Tropp (HMT)~\cite{halko2011finding} for a 40-dimensional covariance estimation task.
HMT is not consistent as $n$ increases, while CSL is. 
This agrees with both our upper (Corollary~\ref{cor:gaussian_upper}) and lower (Proposition~\ref{prop:fixed}) bounds.
}
\label{fig:compare_exp}
\end{figure}

In Figure~\ref{fig:compare_exp}, we plot the spectral norm error of our approach (called CSL) and the algorithm of Halko, Martinsson, and Tropp (called HMT) as a function of the number of samples $n$.
The data is drawn from a 40-dimensional multivariate normal distribution that has covariance matrix with unit spectral norm. 
The main takeaway here is that our approach is consistent as $n$ increases with $d,m$ fixed while the HMT algorithm is not. 
Our approach is consistent because it uses a different random projection for each data vector and averages across these vectors, so it obtains estimates for all directions of the target covariance.
In contrast, the HMT algorithm uses the same (data-dependent) $m$-dimensional projection, and, consequently, only $m \ll d$ directions are ever observed.

\subsection{Guarantees for Subspace Learning}
In the subspace learning problem, the goal is to estimate the principal components of the data, which amounts to the leading eigenvectors of the covariance matrix. 
Specifically, let the covariance matrix $\Sigma$ have eigendecomposition $\sum_{i=1}^d \lambda_i v_i v_i^T$ with $\lambda_1 \ge \ldots \ge \lambda_d \ge 0$ and let $V_k \in \RR^{d \times k}$ be a matrix whose columns are the leading $k$ eigenvectors (i.e. $v_1, \ldots, v_k$).
The goal of subspace learning is to recover a projection matrix $\hat{\Pi} \in \RR^{d\times d}$ that is close to $\Pi_k \triangleq V_k V_k^T$ in spectral norm. 
Recall that the spectral norm difference between two projection matrices is the magnitude of the sine of the largest principal angle between the associated subspaces.

The usual approach to subspace learning is to first construct an estimate $\hat{\Sigma}$ for the covariance and use the leading eigenvectors of $\hat{\Sigma}$ as the subspace estimate. 
In our compressive setting, we form $\hat{\Sigma}$ via Equation~\eqref{eq:estimator}, compute the eigendecomposition $\hat{\Sigma} = \sum_{i=1}^d \hat{\lambda}_i \hat{v}_i\hat{v}_i^T$ and let $\hat{\Pi}$ be the projection onto  $\hat{v}_1, \ldots, \hat{v}_k$. 

To describe our theoretical guarantee, we require the standard notion of signal strength for subspace learning, namely the \emph{eigengap} $\gamma_k = \lambda_k - \lambda_{k+1}$. 
If $\gamma_k$ is large, then the principal subspace is well separated from the remaining directions, whereas if $\gamma_k$ is zero, then the principal subspace is actually unidentifiable. 
Incorporating this signal strength into our estimation error bounds immediately implies the following result on subspace learning from compressive measurements.
For clarity, we present this result in the Gaussian setting. 
\begin{corollary}[Subspace Learning]
\label{cor:subspace_learning}
Let $x_1, \ldots, x_n \sim \Ncal(0, \Sigma)$ and consider the compressive sampling model under the assumptions in Corollary~\ref{cor:gaussian_upper}. 
There exists a universal constant $\kappa > 0$ such that for any $\delta \in (0,1)$, with $d \ge 2$ and $n \ge d \log(1/\delta)$, with probability at least $1-\delta$,
\ifthenelse{\equal{\numcolumns}{one}}{
\begin{align*}
\|\hat{\Pi} - \Pi_k\|_2 &\le \frac{\kappa \|\Sigma\|_2}{\gamma_k} \left(\sqrt{\frac{d^3\log^2(nd/\delta)}{nm^2}} + \frac{d^3\log^2(nd/\delta)}{nm^2} + \sqrt{\frac{d \log(1/\delta)}{n}}\right).
\end{align*}
}{
\begin{align*}
\|\hat{\Pi} - \Pi_k\|_2 &\le \frac{\kappa \|\Sigma\|_2}{\gamma_k} \left(\sqrt{\frac{d^3\log^2(nd/\delta)}{nm^2}} + \frac{d^3\log^2(nd/\delta)}{nm^2}\right.\\
 &\left. \qquad + \sqrt{\frac{d \log(1/\delta)}{n}}\right).
\end{align*}
}
\end{corollary}

This error bound is a consequence of Corollary~\ref{cor:gaussian_upper} followed by the Davis-Kahan theorem~\cite{davis1970rotation} characterizing how perturbing a matrix affects the eigenvectors.
The only other result for this specific problem is our previous work~\cite{krishnamurthy2014subspace}, which, as we mentioned, uses a weaker guarantee for covariance estimation, and hence gives a weaker bound.
Other results for subspace learning in different measurement settings are similar in spirit to the one here~\cite{gonen2014sample}. 

\subsection{Consequences for Distributed Covariance Estimation}
\label{sec:distributed}
In distributed sensor networks, one is often tasked with performing statistical analysis under both measurement and communication constraints. 
In the distributed covariance estimation problem, the data vectors $x_1, \ldots, x_n$ are observed at $n$ sensors $s_1, \ldots, s_n$ (i.e. sensor $s_t$ observes sample $x_t$), and we would like to estimate the covariance structure of the vectors while incurring minimal measurement and communication overhead.
Applications include environmental and atmospheric monitoring, where each dimension of the data vectors is associated with a particular chemical, so that each sensor records the chemical concentrations at a particular location in the environment/atmosphere and the goal is to understand correlations between these concentrations.
Typically, communication is with a \emph{fusion center} that aggregates the measurements from all of the sensors and performs any additional computation.
Our approach provides a low-cost solution to these problems.

In our approach, each sensor $s_t$ makes $m$ compressive measurements of the signal $x_t$, computes the back-projection $\Phi_tx_t$ and sends this to the fusion center.
This approach has measurement cost $O(nm)$ and communication cost $O(nd)$ as $d$-dimensional vectors must be transmitted to the fusion center. 
If the orthonormal bases $A_t$ used for sensing are synchronized with the fusion center before data acquisition, then the sensors can instead transmit $A_t^Tx_t$, which would result in $O(nm)$ communication cost.
As we saw, the error depends on the effective sample size $nm^2/d^2$ so a practitioner can adjust $m$ to trade-off between measurement overhead and statistical accuracy.
One extreme of this trade-off does not compress the signals at all during measurement; this na\"{i}ve approach has $O(nd)$ measurement and communication cost.

The other natural approach uses a single measurement matrix $A \in \RR^{d \times m}$ at all sensors and obtains the observations $A^Tx_t$. 
This protocol has $O(nm)$ measurement cost and either $O(nm)$ or $O(nmd)$ communication cost, depending on whether the shared measurement matrix is synchronized prior to acquisition or not.
However, Proposition~\ref{prop:fixed} below, shows that this approach is not consistent for the covariance estimation problem unless $m=d$, which offers no measurement savings. 
Thus, our approach offers a favorable solution as either measurement or communication cost is reduced over existing approaches. 
Moreover, we precisely quantify the trade-off between measurement and communication overhead on one hand and statistical accuracy on the other hand.

\section{Lower Bounds}
\label{sec:lower}
We now turn to establishing lower bounds for the compressive covariance estimation problem. 
These lower bounds show that, modulo logarithmic factors, our estimator is \emph{rate-optimal}.
This means that our estimator nearly achieves the best performance one could hope for in terms of the problem parameters $n,m$, and $d$. 

We study the \textbf{minimax risk}, which is the worst-case error of the best estimator. 
Specifically, it is the infimum, over all measurable estimators $\hat{\Sigma}$, of the supremum, over all covariance matrices $\Sigma$, of the expected error (in infinity or spectral norm) of the estimator when the data is generated according to a distribution with covariance $\Sigma$. 
This is therefore the distributional setting, and we will subsequently address the distribution-free setting. 
Formally, we are interested in lower bounding
\begin{align*}
\Rcal_n(\Theta) \triangleq \inf_{\hat{\Sigma}} \sup_{\Sigma \in \Theta} \EE_{\overset{x_1^n \sim P_\Sigma}{\Phi_1^n \sim \Ucal_m}}\left[\|\hat{\Sigma}(\{(\Phi_t, \Phi_tx_t)\}_{t=1}^n) - \Sigma\|\right],
\end{align*}
where the norm $\|\cdot\|$ is either the infinity or the spectral norm and the class $\Theta$ is some subset of the semidefinite cone in $d$ dimensions.
Here, the expectation is over the data vectors $x_1, \ldots, x_n$, which are drawn independently and identically from some distribution $P_\Sigma$ with covariance $\Sigma$, and the projection matrices $\Phi_1, \ldots, \Phi_n$, which are drawn uniformly at random from the set of $m$-dimensional projections in $\RR^d$ (We use $\Ucal_m$ to denote this distribution.). 
The estimator $\hat{\Sigma}$ is parameterized by both the projection operators $\Phi_t$ and the observations $\Phi_tx_t$. 
Note that $\Rcal_n(\Theta)$ also depends on both $m$ and $d$, which we leave implicit. 
We use $\Rcal_{n,2}(\Theta)$ to denote the minimax spectral norm risk and $\Rcal_{n,\infty}(\Theta)$ for the $\ell_{\infty}$ version.
Note that the high probability bounds in Theorems~\ref{thm:infinity_upper} and~\ref{thm:spectral_upper_tight} can be translated into upper bounds on this minimax risk in both norms for appropriate classes $\Theta$. 
Here we are interested in lower bounds. 

Let $\Theta(\ell_{\infty}, \eta, d)$ denote the set of $d$-dimensional positive semidefinite matrices with $\ell_{\infty}$-norm upper bounded by $\eta$. 
Our first theorem in this section lower bounds the minimax $\ell_{\infty}$ error when the data is generated according to a zero-mean Gaussian with covariance matrix $\Sigma \in \Theta(\ell_{\infty}, \eta, d)$, which means that we set $P_\Sigma = \Ncal(0, \Sigma)$.
\begin{theorem}[$\ell_{\infty}$ Lower Bound]
\label{thm:infinity_lower}
If $n \ge \frac{1}{5}\frac{d^2\log d}{m^2}$ and $d \ge 2$, then we have
\begin{align*}
\Rcal_{n,\infty}(\Theta(\ell_{\infty}, \eta, d)) \ge \frac{\eta}{14} \sqrt{\frac{d^2\log d}{15nm^2}}.
\end{align*}
\end{theorem}

For the spectral norm lower bound, let $\Theta(\ell_2, \eta, d)$ denote the set of $d$-dimensional positive semidefinite  matrices with spectral norm at most $\eta$. 
The following theorem lower bounds the spectral norm error when the data is generated according to a zero-mean Gaussian with covariance matrix $\Sigma \in \Theta(\ell_2, \eta, d)$. 
\begin{theorem}[Spectral Lower Bound]
\label{thm:spectral_lower}
If $n \ge \frac{1}{576}\frac{d^3}{m^2}$ and $d \ge 10$, then we have
\begin{align*}
\Rcal_{n,2}(\Theta(\ell_{2}, \eta, d)) \ge \frac{\eta}{408}\sqrt{\frac{d^3}{nm^2}}. 
\end{align*}
\end{theorem}

These two bounds hold in the Gaussian setting and should be compared with the bounds in Corollary~\ref{cor:gaussian_upper}. 
While the constants and logarithmic factors disagree, we see that the leading terms in the rates match in their dependence on $n,m$, and $d$.
Thus, our estimator achieves the minimax rate, modulo logarithmic factors. 
Note that in the lower bounds, one should set $\eta = \|\Sigma\|_{\infty}$ or $\|\Sigma\|_2$ respectively, so that the bounds also agree in the dependence on the signal strength parameters.

The proofs of these two results are based on a standard information-theoretic approach to establishing minimax lower bounds. 
The idea is to reduce the estimation problem to a hypothesis testing problem among many well-separated parameters and lower bound the probability of error in testing.
One obtains this lower bound by applying Fano's inequality, which requires upper bounds on the Kullback-Leibler (KL) divergences between the distributions induced by the parameters.
Our main technical result is a strong data-processing inequality (Lemma~\ref{lem:kl_bound}) that shows that compressing the distributions leads to a significant contraction in the KL-divergence.
We prove this result by exploiting rotational invariance of the selected distributions and properties of Beta random variables.
This contraction leads to much sharper lower bounds than the uncompressed setting.

However, these results do not immediately imply distribution-free lower bounds, as the definition of minimax risk includes an expectation over the data.
Since our upper bounds hold in a distribution-free sense, it is also worth asking if we can establish lower bounds in that setting.
This question can be answered in the affirmative, essentially be reverse-engineering the translation from Theorems~\ref{thm:infinity_upper} and~\ref{thm:spectral_upper_tight} to Corollary~\ref{cor:gaussian_upper}. 
Specifically, in the following result, we generate data from Gaussian distributions but then ask for an estimate of the sample covariance. 
At a high level, we argue that if one cannot estimate the population covariance well, for which Theorems~\ref{thm:infinity_lower} and~\ref{thm:spectral_lower} apply, then one cannot hope to estimate the sample covariance either. 
This gives a distribution-free lower bound on the compressive covariance estimation problem. 

Our distribution-free lower bounds involve the same sample-dependent quantities that arise in the corresponding upper bounds.
To that end, define $\Theta(\eta_\infty)$ to be the set of all $n$-sample data sets $X \in \RR^{d \times n}$ with $\|X\|_{\infty}^2 \le \eta_{\infty}$.
Define $\Theta(\eta_{S_1})$ in a similar way but with the constraint $\|\frac{1}{n}\sum_{t=1}^Tx_tx_t^T\|_2\|X\|_{2,\infty}^2 \le \eta_{S_1}$ and $\Theta(\eta_{S_2})$ with the constraint $\frac{1}{n}\sum_{t=1}^n \|x_t\|_2^4 \le \eta_{S_2}$.
These are closely related to the sample-dependent quantities $\|X\|_{\infty}^2, S_1$, and $S_2$ from before.
We prove distribution-free lower bounds under the assumption that the data set $X$ belongs to one of these classes.
\begin{theorem}[Distribution-Free Lower Bounds]
\label{thm:dfree_lb}
There exists positive constants $d_0, \kappa_1, \kappa_2,\kappa_3$ and a function $n_0: \NN \rightarrow \NN$ such that for each $d \ge d_0$ and $n \ge n_0(d)$, we have
\ifthenelse{\equal{\numcolumns}{one}}{
\begin{align}
\inf_{\hat{\Sigma}} \sup_{X \in \Theta(\eta_{\infty})}& \EE_{\Phi_1^n \sim \Ucal}[\|\hat{\Sigma} - \frac{1}{n}\sum_{t=1}^nx_tx_t^T\|_{\infty}] \ge \frac{\kappa_1\eta_\infty}{\log(nd)}\left(\sqrt{\frac{d^2\log(d)}{nm^2}} - \sqrt{\frac{\log(d)}{n}}\right), \label{eq:dfree_infinity_lower}\\
\inf_{\hat{\Sigma}} \sup_{X \in \Theta(\eta_{S_1})}& \EE_{\Phi_1^n \sim \Ucal}[\|\hat{\Sigma} - \frac{1}{n}\sum_{t=1}^nx_tx_t^T\|_2] \ge \kappa_2 \sqrt{\frac{\eta_{S_1}}{\log(nd)}} \left(\sqrt{\frac{d}{nm}} - \sqrt{\frac{1}{n}}\right), \label{eq:dfree_spectral_1}\\
\inf_{\hat{\Sigma}} \sup_{X \in \Theta(\eta_{S_2})}& \EE_{\Phi_1^n \sim \Ucal}[\|\hat{\Sigma} - \frac{1}{n}\sum_{t=1}^nx_tx_t^T\|_2] \ge \kappa_3 \frac{\sqrt{\eta_{S_2}}}{\log(nd)} \left(\sqrt{\frac{d}{nm^2}} - \sqrt{\frac{1}{dn}}\right). \label{eq:dfree_spectral_2}
\end{align}
}{
\begin{align}
\inf_{\hat{\Sigma}} \sup_{X \in \Theta(\eta_{\infty})}& \EE_{\Phi_1^n \sim \Ucal}[\|\hat{\Sigma} - \frac{1}{n}\sum_{t=1}^nx_tx_t^T\|_{\infty}] \ge \notag \\
& \frac{\kappa_1\eta_\infty}{\log(nd)}\left(\sqrt{\frac{d^2\log(d)}{nm^2}} - \sqrt{\frac{\log(d)}{n}}\right), \label{eq:dfree_infinity_lower}\\
\inf_{\hat{\Sigma}} \sup_{X \in \Theta(\eta_{S_1})}& \EE_{\Phi_1^n \sim \Ucal}[\|\hat{\Sigma} - \frac{1}{n}\sum_{t=1}^nx_tx_t^T\|_2] \ge \notag \\
& \kappa_2 \sqrt{\frac{\eta_{S_1}}{\log(nd)}} \left(\sqrt{\frac{d}{nm}} - \sqrt{\frac{1}{n}}\right), \label{eq:dfree_spectral_1}\\
\inf_{\hat{\Sigma}} \sup_{X \in \Theta(\eta_{S_2})}& \EE_{\Phi_1^n \sim \Ucal}[\|\hat{\Sigma} - \frac{1}{n}\sum_{t=1}^nx_tx_t^T\|_2] \ge \notag \\
& \kappa_3 \frac{\sqrt{\eta_{S_2}}}{\log(nd)} \left(\sqrt{\frac{d}{nm^2}} - \sqrt{\frac{1}{dn}}\right). \label{eq:dfree_spectral_2}
\end{align}
}
\end{theorem}
We remark that the conditions on $d$ and $n$ enable us to apply Theorems~\ref{thm:infinity_lower} and~\ref{thm:spectral_lower}, although we have omitted the actual constraints for clarity of presentation.

This theorem is best compared with Theorems~\ref{thm:infinity_upper} and~\ref{thm:spectral_upper_tight}.
Ignoring logarithmic factors, the right hand side of Equation~\eqref{eq:dfree_infinity_lower} is $\tilde{\Omega}\left(\eta_{\infty}\sqrt{\frac{d^2}{nm^2}}\right)$, which matches the leading order term in our distribution-free $\ell_{\infty}$ upper bound (Theorem~\ref{thm:infinity_upper}) when $\frac{d^2}{nm^2} \le 1$.
We prove two distribution-free spectral norm lower bounds to match the two terms in Theorem~\ref{thm:spectral_upper_tight}.
The first bound, Equation~\eqref{eq:dfree_spectral_1}, has leading order term $\tilde{\Omega}\left(\sqrt{\frac{d\eta_{S_1}}{nm}}\right)$ while the second, Equation~\eqref{eq:dfree_spectral_2}, has leading order term $\tilde{\Omega}\left(\sqrt{\frac{d\eta_{S_2}}{nm^2}}\right)$.
These match the terms in Theorem~\ref{thm:spectral_upper_tight} once we replace $S_1$ with $\eta_{S_1}$ and $S_2$ with $\eta_{S_2}$ and provided that $\frac{d}{nm} \le S_1/\|X\|_{2,\infty}^4$.
This last condition, and the condition $\frac{d^2}{nm^2} \le 1$ under which we compare the $\ell_{\infty}$ norm bounds put us in the more interesting regime where non-trivial estimation is possible.
Indeed, if $\frac{d^2}{nm^2} > 1$, Equation~\eqref{eq:dfree_infinity_lower} precludes error smaller than $\tilde{\Omega}(\eta_{\infty})$, but this can be trivially achieved with $\hat{\Sigma} = 0$ since $\|\frac{1}{n}\sum_{t=1}^nx_tx_t^T\|_{\infty} \le \|X\|_\infty^2 \le \eta_{\infty}$. 
Similarly, if $\frac{d}{nm} >  \eta_{S_1}/\|X\|_{2,\infty}^4$, then Equation~\eqref{eq:dfree_spectral_1} precludes error smaller than $\tilde{\Omega}(\eta_{S_1}/\|X\|_{2,\infty}^2)$, and again this can be trivially obtained by the all-zeros estimator by the definition of $\eta_{S_1}$.

Lastly, we consider a different compression scheme where, rather than drawing an independent random projection for each data vector, we use the same random projection on every sample. 
As we have mentioned, this approach has been used in several recent papers to estimate structured covariance matrices~\cite{dasarathy2013sketching,cai2015rop,chen2013exact}.
The following bound shows that this approach is \emph{inconsistent} for the unstructured setting. 
Intuitively, the challenge is that one simply does not observe $d-m$ directions of the covariance matrix, so one cannot hope to estimate the energy in these directions.
This intuition is formalized in the following.
\begin{proposition}[Shared compression operator lower bound]
As long as $m < d$,
\begin{align*}
\inf_{T} \sup_{\stackrel{\Sigma \succeq 0}{\|\Sigma\|_2 \le \eta}} \EE_{\Pi \sim \Ucal_m}\|T(\Pi, \Pi\Sigma\Pi) - \Sigma \|_2 \ge \frac{\eta}{\sqrt{2}}\left(1 - \frac{m}{d}\right)^{1/2},
\end{align*}
so that consistent estimation of $\Sigma$ is impossible with fixed $m$-dimensional projection operator. 
\label{prop:fixed}
\end{proposition}

Notice that in this theorem, the estimator $T$ actually has access to a compressed version of the target covariance $\Sigma$. 
The expectation is over the randomness in the projection, and holds in an asymptotic sense, with no dependence on the number of samples $n$. 
Consequently, we see that consistent recovery of the covariance matrix $\Sigma$ is not possible unless $m = d$, in which case it is trivial. 
This shows that this fixed-compression sampling scheme is not suitable for the unstructured covariance estimation problem. 

In addition to the fact that this is a population level analysis, the proof also departs from traditional minimax lower bound techniques in that we do not use a discretization of the hypothesis space, which in this case is the semidefinite cone. 
Instead, we lower bound the supremum with an expectation over a continuous distribution and use the geometric structure of this distribution to explicitly lower bound the expectation. 
Specifically, we reduce the compressed covariance estimation problem to a compressed vector estimation problem, and we lower bound the expected error for this problem when the vector is distributed uniformly on the unit sphere and the projection is distributed uniformly over the set of $m$-dimensional projections.
To our knowledge, this approach to proving minimax lower bounds is novel, and we believe it will be applicable in other compressed sensing problems.

\section{Proofs}
\label{sec:proofs}
In this section we provide proofs of our main theorems and corollaries.
We begin by introducing some tools that we will use in the proofs, turn next to the upper bounds, and close this section with proofs of the lower bounds.
To maintain readability, the proofs of many lemmas stated in this section are deferred to the appendices. 

\subsection{Preliminary Tools}

We will make extensive use of the properties of the Beta distribution so we collect several facts here.
A random variable $\omega$ supported on $[0,1]$ is said to be Beta distributed with shape parameters $\alpha, \beta > 0$ if it has probability density function $p(\omega) = \frac{\omega^{\alpha-1}(1-\omega)^{\beta-1}}{B(\alpha,\beta)}$ where $B(\cdot, \cdot)$ is the Beta function.
\begin{fact}[Properties of Beta Distribution]
he following facts involving random projections and the Beta distribution hold:
\begin{enumerate}
\item Let $a \sim \chi_{m}^2, b \sim \chi_{d-m}^2$ be independent Chi-squared distributed random variables.
  Then $\frac{a}{a+b} \sim \Beta(\frac{m}{2}, \frac{d-m}{2})$. 
\item Let $\omega \sim \Beta(\frac{m}{2}, \frac{d-m}{2})$.
  Then
\ifthenelse{\equal{\numcolumns}{one}}{
\begin{align*}
\EE[\omega^i] = \prod_{j=1}^i \frac{m+2(j-1)}{d+2(j-1)}, \qquad \textrm{and} \qquad 
\EE[(\omega-\omega^2)] = \frac{m(d-m)}{d(d+2)}.
\end{align*}
}{
\begin{align*}
\EE[\omega^i] & = \prod_{j=1}^i \frac{m+2(j-1)}{d+2(j-1)}, \\
\EE[(\omega-\omega^2)] & = \frac{m(d-m)}{d(d+2)}.
\end{align*}
}
\item If $x \in \RR^d$ and $\Phi \in \RR^{d\times d}$ is a uniformly distributed rank $m$ orthogonal projection, then
\begin{align*}
\Phi x \,{\buildrel d \over =}\, \omega x + \sqrt{\omega - \omega^2}\|x\|W\alpha,
\end{align*}
where $\omega \sim \textrm{Beta}\left(\frac{m}{2}, \frac{d-m}{2}\right)$, $\alpha \in \RR^{d-1}$ distributed uniformly on the unit sphere and independent from $\omega$, and $W \in \RR^{d\times(d-1)}$ is an orthonormal basis for the subspace orthogonal to $x$, i.e. $x^TW = 0$. 
\end{enumerate}
\label{fact:beta}
\end{fact}

For the subspace learning application we will also use the Davis-Kahan theorem, which is a standard result from matrix perturbation theory.
The theorem characterizes how an additive perturbation to a matrix affects its eigenvectors.
\begin{theorem}[Davis-Kahan Theorem~\cite{davis1970rotation}]
\label{thm:davis_kahan}
Let $M,A \in \RR^{d \times d}$ be symmetric matrices with eigenvectors $v_1, \ldots, v_d$ (resp. $u_1, \ldots, u_d$) and eigenvalues $\lambda_1 \ge \ldots \ge \lambda_d$ (resp. $\mu_1 \ge \ldots \ge \mu_d$). 
Define $\delta_i = \min_{j \ne i} |\lambda_i - \lambda_j|$. Then, for an $i \in [d]$,
\begin{align*}
\sin \angle (v_i, u_i) \le \frac{\|M-A\|_2}{\delta_i}.
\end{align*}
\end{theorem}

We will also use several standard concentration inequalities for Gaussian random vectors. 
\begin{proposition}
\label{prop:gaussian_tails}
Let $x_1, \ldots, x_n \sim \Ncal(0, \Sigma)$ in $\RR^d$ where $d \ge 2$ and let $X \in \RR^{d \times n}$ have $t^{\textrm{th}}$ column $x_t$ . 
Then there exists a universal constant $c > 0$ such that for any $\delta \in (0,1)$, the following tail bounds on $X$ hold:
\begin{align*}
\PP\left(\|X\|_{\infty} \le \sqrt{2\|\Sigma\|_{\infty}\log(nd/\delta)}\right) &\ge 1-\delta,\\
\PP\left(\|X\|_{2,\infty} \le \sqrt{2\tr(\Sigma)\log(nd/\delta)}\right) & \ge 1-\delta,\\
\PP\left(\left\|\frac{1}{n}\sum_{t=1}^nX_tX_t^T - \Sigma\right\|_2 \le \|\Sigma\|_2 \sqrt{\frac{cd\log(2/\delta)}{n}}\right) & \ge 1-\delta,\\
\PP\left(\left\|\frac{1}{n}\sum_{t=1}^nX_tX_t^T - \Sigma\right\|_{\infty} \le \|\Sigma\|_{\infty} \sqrt{\frac{\log(2d/\delta)}{n}}\right) & \ge 1-\delta.
\end{align*}
Moreover if $\textrm{rank}(\Sigma) \le k$, we have
\begin{align*}
\PP\left(\|\frac{1}{n}\sum_{t=1}^nX_tX_t^T - \Sigma\|_2 \le \|\Sigma\|_2 \sqrt{\frac{ck\log(2/\delta)}{n}}\right) & \ge 1-\delta.
\end{align*}
\end{proposition}
All four of these are standard. 
The first two follow from Gaussian tail bounds and a union bound over the $n$ vectors.
The third result is based on random matrix theory, and shows that the usual sample covariance matrix is a good estimator for the population in spectral norm~\cite{vershynin2010introduction}.
The fourth bound uses $\chi^2$ tails to give $\ell_{\infty}$ norm bounds on the error of sample covariance matrix. 
Finally, the low-rank bound follows from the proof of the third bound, since there is no error in the $d-k$ directions orthogonal to $\Sigma$. 

\subsection{Proof of Proposition~\ref{prop:unbiasedness}}
\label{sec:proof_unbiased}
It suffices to consider $n = 1$ as the result will follow by linearity of expectation.
Let $x = x_1$, $\Phi = \Phi_1$, and write $\Phi = VV^T$ for $V \in \RR^{d\times m}$ with orthonormal columns. 
Note that $V$ is a random variable.

We first prove the third claim of Fact~\ref{fact:beta}.
To characterize $\|\Phi x\|_2^2$, first note that $\|\Phi x\|_2^2 = \|V^Tx\|_2^2$ since $V$ has orthonormal columns. 
Next, observe that for any rotation matrix $R$ (i.e. matrix with $RR^T = I$) we have $\|V^T x\|_2^2 = \|V^T RR^T x\|_2^2$. 
We draw $R$ conditionally on $V$ from the uniform distribution over all rotation matrices such that $V^T R = (I_m, 0_{m,d-m})$, i.e. the $m \times d$ matrix with $I_m$ in the first $m$ columns and zero everywhere else. 
Now since $\Phi$ and hence $V$ are drawn uniformly at random, we can equivalently draw $R$ uniformly at random from the set of all rotation matrices, and then draw $V$ conditional on $R$ such that $V^T R = (I_m, 0_{m,d-m})$.
Thus, $R^Tx$ is a uniform at random vector with norm $\|x\|_2$ and therefore $ \| \Phi R R^T x \|_2^2 \equalsd \|(I_m, 0_{m,d-m}) u\|_2^2 \times \|x\|_2^2 \equalsd \|\Pi_m u\|_2^2 \times \|x\|_2^2$ where $u$ is a uniform at random unit vector and $\Pi_m$ projects onto the first $m$ standard basis elements. 
This technique of viewing the projection as onto the standard basis and randomizing the vector $u$ is common in the analysis of random projection methods~\cite{dasgupta2003elementary}.
Recall that $x$ is not a random variable. 

We now characterize $\|\Pi_m u\|_2^2$. 
If $z_1,\ldots, z_d \sim \Ncal(0,1)$, we have $u \equalsd (\frac{z_1}{\sqrt{\sum_j z_j^2}}, \ldots, \frac{z_d}{\sqrt{\sum_j z_j^2}})$.
This identity implies that
\begin{align*}
\|\Pi_m u\|_2^2 \equalsd \sum_{i=1}^m \frac{z_i^2}{\sum_{j=1}^d z_j^2} \sim \Beta\left(\frac{m}{2},\frac{d-m}{2}\right)
\end{align*}
using Claim 1 of Fact~\ref{fact:beta} and the relationship between Gaussian and $\chi^2$ random variables. 
Thus, we have that $\|\Phi x\|_2^2 \equalsd \omega \|x\|_2^2$ where $\omega \sim \Beta\left(\frac{m}{2}, \frac{d-m}{2}\right)$.

This last fact implies that $\cos \angle(x, \Phi x) \equalsd \sqrt{\omega}$, or, equivalently, that the magnitude of $\Phi x$ in the direction of $x$ is $\|\Phi x\|_2 \cos \angle(x, \Phi x) \equalsd \omega\|x\|_2$.
By the Pythagorean Theorem, the magnitude of $\Phi x$ in the orthogonal direction must therefore be $\|x\|_2 \sqrt{\omega - \omega^2}$, and this direction is chosen uniformly at random and independently from $\omega$, subject to being orthogonal to $x$.
This gives the identity $\Phi x \equalsd \omega x + \sqrt{\omega - \omega^2}\|x\|W\alpha$, which is Claim 3 of Fact~\ref{fact:beta}.

Claim 3 of Fact~\ref{fact:beta} means that
\ifthenelse{\equal{\numcolumns}{one}}{
\begin{align*}
\Phi xx^T \Phi^T &\equalsd \omega^2 xx^T + (\omega - \omega^2) \|x\|^2 W\alpha \alpha^T W^T + \omega\sqrt{\omega - \omega^2}\|x\|\left(x\alpha^TW^T + W\alpha x^T\right).
\end{align*}
}{
\begin{align*}
\Phi xx^T \Phi^T &\equalsd \omega^2 xx^T + (\omega - \omega^2) \|x\|^2 W\alpha \alpha^T W^T\\
& + \omega\sqrt{\omega - \omega^2}\|x\|\left(x\alpha^TW^T + W\alpha x^T\right).
\end{align*}
}
By linearity of expectation we can analyze each term individually.
By Fact~\ref{fact:beta} and the distribution of $\alpha$, we know that $\EE \omega = \frac{m}{d}$, $\EE\omega^2 = \frac{m(m+2)}{d(d+2)}$, $\EE\alpha = 0$, and $\EE \alpha \alpha^T = \frac{1}{d-1} I_{d-1}$ since $\alpha$ is distributed uniformly on the $d-1$ dimensional sphere. 
This means
\ifthenelse{\equal{\numcolumns}{one}}{
\begin{align*}
\EE \Phi x x^T\Phi = &\frac{m (m+2)}{d (d+2)} xx^T + \frac{m (d-m)}{d (d+2)} \|x\|_2^2 \EE (W \alpha \alpha^T W^T) + \EE(\omega \sqrt{\omega - \omega^2}) \|x\| \left( x \EE \alpha^T W^T + W\EE \alpha x^T\right)\\
= &\frac{m (m+2)}{d (d+2)} xx^T + \frac{m(d-m)}{d(d+2)(d-1)} \|x\|_2^2 WW^T\\
= & \frac{m (m+2)}{d (d+2)} xx^T + \frac{m(d-m)}{d(d+2)(d-1)} \|x\|_2^2 \left(I - \frac{xx^T}{\|x\|_2^2}\right)\\
= &\frac{m (md+d-2)}{d(d+2)(d-1)} xx^T + \frac{m (d-m)}{d(d+2)(d-1)} \|x\|_2^2 I.
\end{align*}
}{
\begin{align*}
& \EE \Phi x x^T\Phi\\
& = \frac{m (m+2)}{d (d+2)} xx^T + \frac{m (d-m)}{d (d+2)} \|x\|_2^2 \EE (W \alpha \alpha^T W^T)\\
 & + \EE(\omega \sqrt{\omega - \omega^2}) \|x\| \left( x \EE \alpha^T W^T + W\EE \alpha x^T\right)\\
& = \frac{m (m+2)}{d (d+2)} xx^T + \frac{m(d-m)}{d(d+2)(d-1)} \|x\|_2^2 WW^T\\
& = \frac{m (m+2)}{d (d+2)} xx^T + \frac{m(d-m)}{d(d+2)(d-1)} \|x\|_2^2 \left(I - \frac{xx^T}{\|x\|_2^2}\right)\\
& = \frac{m (md+d-2)}{d(d+2)(d-1)} xx^T + \frac{m (d-m)}{d(d+2)(d-1)} \|x\|_2^2 I.
\end{align*}
}
Note that $\|x\|_2^2 = \tr(xx^T)$.
Proposition~\ref{prop:unbiasedness} then follows by linearity of expectation, using the same expansion for all of the $n$ samples and rescaling by $d^2/m^2$.

\subsection{Upper Bounds}
Recall that $\hat{\Sigma}_1 = \frac{d^2}{nm^2}\sum_{t=1}^n \Phi_t x_t x_t^T \Phi_t$ is the observed covariance.
Define
\begin{align*}
\bar{\Sigma} \triangleq \frac{d(dm+d-2)}{m(d+2)(d-1)}\Sigma + \frac{d(d-m)}{m(d+2)(d-1)} \tr(\Sigma)I,
\end{align*}
which is the expectation of $\hat{\Sigma}_1$ from Proposition~\ref{prop:unbiasedness}.
The proofs of both infinity and spectral norm bounds follow by arguing that $\hat{\Sigma}_1$ is close to $\bar{\Sigma}$ and then using this fact to relate our estimator $\hat{\Sigma}$ to the estimand of interest $\Sigma$. 

\medskip
\noindent\textbf{$\ell_{\infty}$-norm Bound:}
The main ingredient of the $\ell_{\infty}$ bound is an intermediary deviation bound on quadratic forms. 
The proof is deferred to the appendix.
\begin{lemma}[Quadratic-Form Deviation Bound]
\label{lem:quadratic_form}
Let $d \ge 2$.
For any unit vector $u \in \RR^d$, define $b \triangleq \max_{t \in [n]} |x_t^Tu|$ and $c \triangleq \max_{t \in [n]} \sqrt{\|x_t\|_2^2 - (x_t^Tu)^2}$. 
For any $\delta \in (0,1)$ with $\delta \le n/e$ and $\log(1/\delta) \le \frac{n}{9600} \frac{d(d+3)^2}{(d-1)^2(d+1)}$, with probability at least $1-4\delta$, we have
\ifthenelse{\equal{\numcolumns}{one}}{
\begin{align*}
\left|u^T(\hat{\Sigma}_1 - \bar{\Sigma})u\right| & \le \sqrt{\frac{d^2 \log(2/\delta)}{nm^2}} \frac{8c^2}{d} + \sqrt{\frac{d^2 \log(2/\delta)}{nm^2}} \left[ \sqrt{18 \log^2(n/\delta)} \left(b^2 + \frac{16c^2}{d}\right)\right]\\
+ & \frac{4d^2\log(2/\delta)}{nm^2} \left(b^2 + \frac{c^2\log(n/\delta)}{d}\right).
\end{align*}
}{
\begin{align*}
& \left|u^T(\hat{\Sigma}_1 - \bar{\Sigma})u\right| \le \sqrt{\frac{d^2 \log(2/\delta)}{nm^2}} \frac{8c^2}{d}\\
& + \sqrt{\frac{d^2 \log(2/\delta)}{nm^2}} \left[ \sqrt{18 \log^2(n/\delta)} \left(b^2 + \frac{16c^2}{d}\right)\right]\\
& + \frac{4d^2\log(2/\delta)}{nm^2} \left(b^2 + \frac{c^2\log(n/\delta)}{d}\right).
\end{align*}
}
\end{lemma}

We are now able to prove Theorem~\ref{thm:infinity_upper}.
Taking a union bound over all vectors $v_i = e_i$ and $u_{ij} = \frac{e_i + e_j}{\sqrt{2}}$ for $i\ne j \in [d]$ enables us to bound the infinity norm since, for any matrix $M$,
\ifthenelse{\equal{\numcolumns}{one}}{
\begin{align*}
|M_{ij}| \le |M_{ii}|/2 + |M_{jj}|/2 + |M_{ii}/2+M_{jj}/2 + M_{ij}|= \frac{1}{2} |v_i^TMv_i| + \frac{1}{2}|v_j^TMv_j| + |u_{ij}^TMu_{ij}|.
\end{align*}
}{
\begin{align*}
|M_{ij}| &\le |M_{ii}|/2 + |M_{jj}|/2 + |M_{ii}/2+M_{jj}/2 + M_{ij}|\\
& = \frac{1}{2} |v_i^TMv_i| + \frac{1}{2}|v_j^TMv_j| + |u_{ij}^TMu_{ij}|.
\end{align*}
}
Thus, the infinity norm is bounded by twice the maximal quadratic form
among these $d(d+1)/2 \le d^2$ vectors.
These quadratic forms can be controlled with
Lemma~\ref{lem:quadratic_form}, where for each application, we use the
upper bounds $b = \max_{t \in [n]}|x_t^Tu| \le \sqrt{2}
\|X\|_{\infty}$, which holds for our choices $v_i$ and $u_{ij}$ and
$c = \max_{t \in [n]}\sqrt{\|x_t\|_2^2 - (x_t^Tu)^2} \le
\|X\|_{2,\infty}$, which holds for any unit vector $u$.
Via a union bound we have that, with probability $1 - \delta$,
\ifthenelse{\equal{\numcolumns}{one}}{
\begin{align*}
\|\hat{\Sigma}_1-\bar{\Sigma}\|_{\infty} &\le  \kappa_1 \sqrt{\frac{d^2\log(d/\delta)}{nm^2}}\left(\sqrt{\log^2(nd/\delta)}\left(\|X\|_{\infty}^2 + \frac{\|X\|_{2,\infty}^2}{d}\right)\right) 
+ \kappa_2 \frac{d^2\log(d/\delta)}{nm^2}\left(\|X\|_{\infty}^2 + \frac{\|X\|_{2,\infty}^2\log(nd/\delta)}{d}\right)\\
& \le \kappa_1 \|X\|_{\infty}^2 \sqrt{\frac{d^2\log^3(nd/\delta)}{nm^2}} + \kappa_2 \|X\|_{\infty}^2 \frac{d^2\log^2(nd/\delta)}{nm^2}.
\end{align*}
}{
\begin{align*}
& \|\hat{\Sigma}_1-\bar{\Sigma}\|_{\infty} \le \\
&\kappa_1 \sqrt{\frac{d^2\log(d/\delta)}{nm^2}}\left(\sqrt{\log^2(nd/\delta)}\left(\|X\|_{\infty}^2 + \frac{\|X\|_{2,\infty}^2}{d}\right)\right) \\
& + \kappa_2 \frac{d^2\log(d/\delta)}{nm^2}\left(\|X\|_{\infty}^2 + \frac{\|X\|_{2,\infty}^2\log(nd/\delta)}{d}\right)\\
& \le \kappa_1 \|X\|_{\infty}^2 \sqrt{\frac{d^2\log^3(nd/\delta)}{nm^2}} + \kappa_2 \|X\|_{\infty}^2 \frac{d^2\log^2(nd/\delta)}{nm^2}.
\end{align*}
}
for constants $\kappa_1, \kappa_2 > 0$ and using the fact that $\|X\|_{2,\infty}^2 \le d\|X\|_{\infty}^2$. 

Using the relationship between $\hat{\Sigma}$ and $\hat{\Sigma}_1$ (Equation~\eqref{eq:estimator}) and the equivalent relationship between $\Sigma$ and $\bar{\Sigma}$ (Equation~\eqref{eq:unbiased_check}), we then have
\ifthenelse{\equal{\numcolumns}{one}}{
\begin{align*}
& \|\hat{\Sigma} - \Sigma\|_{\infty} \le \frac{m(d+2)(d-1)}{d(dm+d-2)}\|\hat{\Sigma}_1 - \bar{\Sigma}\|_{\infty} + \frac{m(d-m)}{dm+d-2} \|\hat{\Sigma}_1 - \bar{\Sigma}\|_{\infty}  \le 2 \|\hat{\Sigma}_1 - \bar{\Sigma}\|_{\infty},
\end{align*}
}{
\begin{align*}
& \|\hat{\Sigma} - \Sigma\|_{\infty} \le \\
& \frac{m(d+2)(d-1)}{d(dm+d-2)}\|\hat{\Sigma}_1 - \bar{\Sigma}\|_{\infty} + \frac{m(d-m)}{dm+d-2} \|\hat{\Sigma}_1 - \bar{\Sigma}\|_{\infty} \\
& \le 2 \|\hat{\Sigma}_1 - \bar{\Sigma}\|_{\infty},
\end{align*}
}
which uses the bound $\tr(\hat{\Sigma}_1 - \bar{\Sigma}) \le d\|\hat{\Sigma}_1 - \bar{\Sigma}\|_{\infty}$ and holds provided that $d \ge 2$. 
Note that the same relationship holds for the spectral norm, which we will use in the next proof.
Plugging in our upper bound on $\|\hat{\Sigma}_1 - \bar{\Sigma}\|_{\infty}$ completes the proof. 

\medskip
\noindent \textbf{Spectral-norm Bound:}
This proof is an application of a particular version of the Matrix Bernstein inequality that applies to unbounded random variables that exhibit sub-exponential tail decay. 
The result is due to De La Pe\~{n}a and Gin\'{e} (Lemma 4.1.9 in~\cite{delapena2012decoupling}), but we use a version available in Tropp's monograph (Theorem 6.2 in~\cite{tropp2011user}).
For completeness, we reproduce the result as Theorem~\ref{thm:matrix_bernstein_subexp} in the appendix. 

We first decompose the unnormalized sum $\tfrac{m^2}{d^2}\hat{\Sigma}_1$.
Using Fact~\ref{fact:beta}, we have
\begin{align*}
\ifthenelse{\equal{\numcolumns}{one}}{
\frac{m^2}{d^2}\hat{\Sigma}_1 &\equalsd \frac{1}{n}\sum_{t=1}^n\left(\omega_tx_t + \sqrt{\omega_t - \omega_t^2}\|x_t\|_2W_t\alpha_t\right)\times\left(\omega_tx_t + \sqrt{\omega_t - \omega_t^2}\|x_t\|_2W_t\alpha_t\right)^T\\
& = \frac{1}{n}\sum_{t=1}^n\left(\omega_t^2x_tx_t^T + (\omega_t - \omega_t^2)\|x_t\|_2^2W_t\alpha_t\alpha_t^TW_t^T + \omega_t\sqrt{\omega_t - \omega_t^2}\|x_t\|_2\left(x_t\alpha_t^TW_t^T + W_t\alpha_tx_t^T\right)\right)\\
& = \frac{1}{n}\sum_{t=1}^n \left(Y_{1,t} + Y_{2,t} + Y_{3,t}\right),
}{
\frac{m^2}{d^2}\hat{\Sigma}_1 &\equalsd \frac{1}{n}\sum_{t=1}^n\left(\omega_tx_t + \sqrt{\omega_t - \omega_t^2}\|x_t\|_2W_t\alpha_t\right)\\
&\times\left(\omega_tx_t + \sqrt{\omega_t - \omega_t^2}\|x_t\|_2W_t\alpha_t\right)^T\\
& = \frac{1}{n}\sum_{t=1}^n\left(\omega_t^2x_tx_t^T + (\omega_t - \omega_t^2)\|x_t\|_2^2W_t\alpha_t\alpha_t^TW_t^T \right.\\
&\left. + \omega_t\sqrt{\omega_t - \omega_t^2}\|x_t\|_2\left(x_t\alpha_t^TW_t^T + W_t\alpha_tx_t^T\right)\right)\\
& = \frac{1}{n}\sum_{t=1}^n \left(Y_{1,t} + Y_{2,t} + Y_{3,t}\right),
}
\end{align*}
where we have defined
\begin{align*}
Y_{1,t} &\triangleq \omega_t^2 x_tx_t^T,\\
Y_{2,t} &\triangleq  (\omega_t - \omega_t^2)\|x_t\|_2^2W_t \alpha_t \alpha_t^TW_t^T,\\
Y_{3,t} &\triangleq \omega_t\sqrt{\omega_t - \omega_t^2} \|x_t\|_2\left(x_t\alpha_t^TW_t^T + W_t\alpha_t x_t^T\right).
\end{align*}
By linearity of expectation and the triangle inequality,
\begin{align*}
\|\frac{m^2}{d^2}\hat{\Sigma}_1 - \frac{m^2}{d^2}\bar{\Sigma}\|_2 \le \sum_{k=1}^3 \|\frac{1}{n}\sum_{t=1}^nY_{k,t} - \EE Y_{k,t}\|_2.
\end{align*}
The result follows from high probability bounds on the three terms on the right hand side, coupled with a union bound.

To apply the Subexponential Matrix Bernstein inequality, we need to control all moments of these random matrices by particular functions of their variance. 
This is the content of Lemma~\ref{lem:spectral_tail_conditions}, where we show that for each $k \in [3], t \in [n]$, and any integer $p \ge 2$,
\begin{align}
\EE (Y_{k,t} - \EE Y_{k,t})^p \preceq \frac{p!}{2}R_k^{p-2}A_{k,t}^2,
\label{eq:matrix_moment_bound}
\end{align}
for particular scalars $R_k$ and matrices $A_{k,t}$.
$R_k$ is independent of $t$ and serves as range-like term while the $A_{k,t}^2$ matrix appears in the bound through $\sigma_k^2 \triangleq \left\|\sum_{t=1}^n A_{k,t}^2 \right\|$ and acts as the variance term.

\begin{lemma}
\label{lem:spectral_tail_conditions}
Let $S_{1,t} \triangleq \|x_t\|_2^2x_tx_t^T$ and $S_{2,t} \triangleq \|x_t\|_2^4I$.
For integers $p \ge 2$, $k \in [3]$, and $t \in [n]$, Equation~\eqref{eq:matrix_moment_bound} holds with
\begin{align*}
A^2_{1,t} &= 1680 \frac{m^4}{d^4} S_{1,t}, &\qquad R_1 &= 28 \frac{m}{d}\|X\|_{2,\infty}^2,\\
A^2_{2,t} &= 32 \frac{m^2}{d^3}S_{2,t}, &\qquad R_2 &= 16 \frac{m}{d}\|X\|_{2,\infty}^2,\\
A^2_{3,t} & = 15\frac{m^3}{d^3}\left(\frac{2}{d}S_{2,t} + S_{1,t}\right), &\qquad R_3 &= 10\frac{m}{d}\|X\|_{2,\infty}^2.
\end{align*}
\end{lemma}

We now use the definitions
\begin{align*}
\ifthenelse{\equal{\numcolumns}{one}}{
S_1 \triangleq \left\|\frac{1}{n}\sum_{t=1}^n S_{1,t}\right\| = \left\|\frac{1}{n}\sum_{t=1}^n\|x_t\|_2^2x_tx_t^T\right\|_2, \qquad \textrm{and} \qquad 
S_2 \triangleq \left\|\frac{1}{n}\sum_{t=1}^nS_{2,t}\right\| = \frac{1}{n}\sum_{t=1}^n \|x_t\|_2^4.
}{
S_1 &\triangleq \left\|\frac{1}{n}\sum_{t=1}^n S_{1,t}\right\| = \left\|\frac{1}{n}\sum_{t=1}^n\|x_t\|_2^2x_tx_t^T\right\|_2, \\
S_2 &\triangleq \left\|\frac{1}{n}\sum_{t=1}^nS_{2,t}\right\| = \frac{1}{n}\sum_{t=1}^n \|x_t\|_2^4.
}
\end{align*}
With these, the Subexponential Bernstein inequality applied to all three terms reveals that, with probability at least $1-3\delta$,
\begin{align*}
\ifthenelse{\equal{\numcolumns}{one}}{
 \frac{m^2}{d^2}\|\hat{\Sigma}_1 - \bar{\Sigma}\|_2 &\le \frac{1}{n}\left[ \sum_{i=1}^3 \sqrt{2\sigma_i^2\log(d/\delta)} + 2 R_i \log(d/\delta)\right]\\
\le &\left( 48\sqrt{\frac{m^3}{d^3}S_1} + 12\sqrt{\frac{m^2}{d^3}S_2}\right)\sqrt{\frac{2}{n}\log(d/\delta)} + \frac{108 m \|X\|_{2,\infty}^2}{nd}\log(d/\delta).
}{
 \frac{m^2}{d^2}\|\hat{\Sigma}_1 - \bar{\Sigma}\|_2 &\le \frac{1}{n}\left[ \sum_{i=1}^3 \sqrt{2\sigma_i^2\log(d/\delta)} + 2 R_i \log(d/\delta)\right]\\
\le &\left( 48\sqrt{\frac{m^3}{d^3}S_1} + 12\sqrt{\frac{m^2}{d^3}S_2}\right)\sqrt{\frac{2}{n}\log(d/\delta)}\\
 & + \frac{108 m \|X\|_{2,\infty}^2}{nd}\log(d/\delta).
}
\end{align*}
Recall that $\sigma_k^2 = \|\sum_{t=1}^n A_{k,t}^2\|$.
Adjusting for the normalization gives
\begin{align*}
\ifthenelse{\equal{\numcolumns}{one}}{
\|\hat{\Sigma}_1 - \bar{\Sigma}\|_2 &\le \left(48 \sqrt{\frac{d}{m}S_1} + 12 \sqrt{\frac{d}{m^2}S_2}\right)\sqrt{\frac{2}{n}\log(d/\delta)} + \frac{108d\|X\|_{2,\infty}^2}{nm}\log(d/\delta).
}{
\|\hat{\Sigma}_1 - \bar{\Sigma}\|_2 &\le \left(48 \sqrt{\frac{d}{m}S_1} + 12 \sqrt{\frac{d}{m^2}S_2}\right)\sqrt{\frac{2}{n}\log(d/\delta)}\\
 & + \frac{108d\|X\|_{2,\infty}^2}{nm}\log(d/\delta).
}
\end{align*}
Finally, the bound $\|\hat{\Sigma} - \Sigma\|_2 \le 2\|\hat{\Sigma}_1 - \bar{\Sigma}\|_2$ proves the theorem.

\medskip
\noindent \textbf{Proof of Corollary~\ref{cor:gaussian_upper}:}
The corollary follows from Theorems~\ref{thm:infinity_upper} and~\ref{thm:spectral_upper_tight} along with the Gaussian tail bounds in Proposition~\ref{prop:gaussian_tails}.
For both bounds, we first use the triangle inequality to obtain
\begin{align*}
\|\hat{\Sigma} - \Sigma\| \le \|\hat{\Sigma} - \frac{1}{n}\sum_{t=1}^nX_tX_t^T\| + \|\frac{1}{n}\sum_{t=1}^n X_tX_t^T - \Sigma\|.
\end{align*}
Then, for the $\ell_{\infty}$-norm bound, we use the first deviation bound in Proposition~\ref{prop:gaussian_tails} to bound $\|X\|_\infty$ in the application of Theorem~\ref{thm:infinity_upper} and the fourth to bound $\|\frac{1}{n}\sum_{t=1}^n X_tX_t^T - \Sigma\|_{\infty}$.

For the spectral norm bound, we express $S_1 \le \|X\|_{2,\infty}^2 \|\frac{1}{n}\sum_{t=1}^nX_tX_t^T\|_2$ and $S_2 \le d\|X\|_{2,\infty}^2 \|\frac{1}{n}\sum_{t=1}^nX_tX_t^T\|_2$ and then use the second inequality in Proposition~\ref{prop:gaussian_tails} to control the $\|X\|_{2,\infty}^2$ term.
For the sample covariance term, we use the third inequality, which, provided $n \ge cd\log(2/\delta)$, ensures that, with probability at least $1-\delta$,
\begin{align*}
\ifthenelse{\equal{\numcolumns}{one}}{
\|\frac{1}{n}\sum_{t=1}^nX_tX_t^T\|_2 \le \|\frac{1}{n}\sum_{t=1}^nX_tX_t^T- \Sigma\|_2 + \|\Sigma\|_2 \le \|\Sigma\|_2\left(\sqrt{\frac{cd\log(2/\delta)}{n}} + 1\right) \le 2\|\Sigma\|_2.
}{
\|\frac{1}{n}\sum_{t=1}^nX_tX_t^T\|_2 & \le \|\frac{1}{n}\sum_{t=1}^nX_tX_t^T- \Sigma\|_2 + \|\Sigma\|_2 \\
& \le \|\Sigma\|_2\left(\sqrt{\frac{cd\log(2/\delta)}{n}} + 1\right) \le 2\|\Sigma\|_2.
}
\end{align*}
This gives bounds on $S_1$ and $S_2$ of order $O(\|\Sigma\|^2_2 d\log(nd/\delta))$ and $O(\|\Sigma\|_2^2 d^2\log(nd/\delta))$, which we plug into Theorem~\ref{thm:spectral_upper_tight}.
Finally we use the third inequality in Proposition~\ref{prop:gaussian_tails} to control $\|\frac{1}{n}\sum_{t=1}^n X_tX_t^T - \Sigma\|_2$.

\medskip
\noindent \textbf{Proof of Corollary~\ref{cor:low_rank_adapt}:}
The proof is based on a lemma of Achlioptas and McSherry~\cite{achlioptas2007fast} that controls the spectral norm of the matrix $\hat{\Sigma}_k$ in terms of only the $k$ dominant noise directions. 
Specializing their lemma to our setting shows that
\begin{align*}
\|\hat{\Sigma}_k - \Sigma\|_2 \le 2 \|\hat{\Sigma} - \Sigma\|_2,
\end{align*}
which follows since $\Sigma$ is rank $k$. 
To conclude the proof, we must control the quantities $S_1$ and $S_2$ in Theorem~\ref{thm:spectral_upper_tight}.
As before, we use the bound $S_1 \le \|X\|_{2,\infty}^2 \|\Sigma\|_2$, but now it is better to use $S_2 \le \|X\|_{2,\infty}^4$. 
Proposition~\ref{prop:gaussian_tails} shows that with probability at least $1-\delta$, we have the bound
\begin{align*}
\|X\|_{2,\infty}^2 \le 2 \tr(\Sigma) \log(nd/\delta) \le 2k \|\Sigma\| \log(nd/\delta).
\end{align*}
The proof then concludes by applying the triangle inequality as in the proof of Corollary~\ref{cor:gaussian_upper}. We use this expression in the upper bounds for both $S_1$ and $S_2$ and apply Theorem~\ref{thm:spectral_upper_tight} to bound the deviation between the estimator and the sample covariance, and then we use Proposition~\ref{prop:gaussian_tails} to bound the deviation between the sample and population covariances. 

\medskip
\noindent \textbf{Proof of Corollary~\ref{cor:subspace_learning}:}
Corollary~\ref{cor:subspace_learning} follows immediately from the spectral norm bound in Corollary~\ref{cor:gaussian_upper} and the Davis-Kahan Theorem (Theorem~\ref{thm:davis_kahan}) which introduces the eigengap $\gamma_k$. 

\subsection{Lower Bounds}

Our lower bounds employ a well-known argument based on Fano's
inequality.  In particular, we will use the following result
(See~\cite{tsybakov2009introduction}).  For the result, we use the
notion of a semi-distance $\rho$, which satisfies all the properties of a
metric but need not satisfy the identifiability property that $\rho(x,y) = 0$
implies $x=y$.
\begin{theorem}
\label{thm:tsybakov_multi}
Let $\Theta$ be a parameter space equipped with a semi-distance $\rho: \Theta \times \Theta \to [0,\infty)$. 
Assume that $M \ge 2$ and that $\Theta$ contains elements $\theta_0, \theta_1, \ldots, \theta_M$ associated with probability measures $\PP_0, \ldots, \PP_M$ such that:
\begin{enumerate}
\item $\rho(\theta_i, \theta_j) \ge 2s > 0$ for all $0 \le i < j \le M$.
\item $\PP_j$ is absolutely continuous with respect to $\PP_0$ for all $j \in [M]$, and
\begin{align*}
\frac{1}{M}\sum_{j=1}^M KL(\PP_j || \PP_0) \le \alpha \log M,
\end{align*}
with $0 < \alpha < 1/8$.
\end{enumerate}
Then
\begin{align*}
\inf_{\hat{\theta}} \sup_{\theta \in \Theta} \PP_{\theta}\left[ \rho(\hat{\theta}, \theta) \ge s\right] \ge \frac{\sqrt{M}}{1+\sqrt{M}}\left(1 - 2\alpha - \sqrt{\frac{2\alpha}{\log M}}\right).
\end{align*}
\end{theorem}

To apply the theorem we need to control the Kullback-Leibler divergence between these distributions.
The following lemma, proved in the appendix, enables a sharp KL-divergence bound. 
\begin{lemma}[KL-divergence bound]
\label{lem:kl_bound}
Let $\PP_0$ be a distribution on $(z,U)$ where $U \in \RR^{d\times m}$ is an orthonormal basis for a uniform-at-random $m$-dimensional subspace, $x \sim \Ncal(0, \eta I_d)$ and $z = U^Tx$. 
Let $\PP_1$ be the same distribution but where $x \sim \Ncal(0, \eta I_d - \gamma vv^T)$ for any unit vector $v \in \RR^d$ and any $\gamma \in \RR$ such that $\gamma \le \frac{1-e}{e}\eta$.
Then the product measures $\PP_1^n, \PP_0^n$ satisfy
\begin{align}
KL(\PP_1^n || \PP_0^n) \le \frac{3}{2}\frac{\gamma^2}{\eta^2}\frac{nm^2}{d^2}.
\end{align}
\end{lemma}

The lemma demonstrates that our compression model results in a contraction in the Kullback-Leibler divergence between Gaussian distributions. 
As the KL-divergence between the two Gaussians is $\Theta(\frac{\gamma^2}{\eta^2}n)$, this contraction is by a multiplicative factor of $m^2/d^2$.
This is known in the literature as a \emph{strong data-processing inequality}~\cite{anantharam2013maximal,duchi2013local,raginsky2014strong}, and it allows us to easily adapt existing lower bound constructions to our setting.
Notice also that the two covariances matrices used in the lemma are separated in both spectral and $\ell_{\infty}$-norm by $\gamma$. The lemma shows how these these norms can be related to the KL-divergence. 

We will also need a packing of the $d$-dimensional Euclidean sphere in the semi-distance $\rho(u,v) = \|uu^T - vv^T\|_2$, which is the spectral norm between the corresponding rank one matrices and is easily verified to satisfy the semi-distance properties.
Asymptotic results of this form exist in the approximation theory literature, where one classical result is the Chabauty-Shannon-Wyner theorem~\cite{ericson2001codes}.
For our purposes the following non-asymptotic statement, which is weaker than the results in the literature, will suffice.
This lemma is proved in the appendix using the probabilistic method.
\begin{lemma}[Packing number lower bound]
If $d \ge 6$, then there exists a set of unit vectors $\{v_j\}_{j=1}^M$ in $\RR^d$ of size $M = \lceil \exp(-1/8)2^{d/8}\rceil $ such that $\|v_iv_i^T - v_jv_j^T\|_2 \ge 1/2$ for all $i \ne j$. 
\label{lem:packing}
\end{lemma}

\medskip
\noindent \textbf{Proof of Theorem~\ref{thm:infinity_lower}:}
The goal of this proof is to apply Theorem~\ref{thm:tsybakov_multi} with the $\ell_{\infty}$ norm as the semi-distance and with a set of $M=d+1$ distributions. 
Define $\Sigma_0 = \eta I$ and $\Sigma_j = \eta I - \gamma e_je_j^T$ for each $j \in [d]$, where recall that $e_j$ is the $j^{\textrm{th}}$ standard basis vector.
The first distribution, $P^n_0$, has the $n$ samples drawn iid from $\Ncal(0, \Sigma_0)$, while, for each $j \in [d]$, the distribution $P^n_j$ has the $n$ samples drawn from $\Ncal(0, \Sigma_j)$. 
By the triangle inequality, when $j \ne k$, we always have $\rho(\Sigma_j, \Sigma_k) = \|\Sigma_j - \Sigma_k\|_{\infty} \geq \gamma$ so the first condition of Theorem~\ref{thm:tsybakov_multi} is satisfied with $s = \gamma/2$. 
Secondly, the infinity norm bound on all covariance matrices is $\eta$ and, to apply Lemma~\ref{lem:kl_bound}, we require $\gamma \le \frac{e-1}{e}\eta$ which also ensures positive semidefiniteness. 
Lastly, by Lemma~\ref{lem:kl_bound}, we have
\begin{align*}
\frac{1}{d}\sum_{j=1}^d KL(P^n_j || P^n_0) \le \frac{3}{2}\frac{\gamma^2}{\eta^2}\frac{nm^2}{d^2},
\end{align*}
which means that we can set $\gamma = \eta \sqrt{\dfrac{2\alpha}{3}\dfrac{d^2\log d}{nm^2}}$. 
The constraint on $\gamma$ is satisfied provided that
\begin{align*}
\sqrt{\dfrac{2\alpha}{3}\dfrac{d^2\log d}{nm^2}} \le \frac{e-1}{e},
\end{align*}
which will lead to the condition involving $n$, $m$ and $d$, once we set $\alpha$. 
Theorem~\ref{thm:tsybakov_multi} now states that
\ifthenelse{\equal{\numcolumns}{one}}{
\begin{align*}
\inf_{\hat{\Sigma}} \sup_{\Sigma} & \PP_{\Sigma}\left[\|\hat{\Sigma} - \Sigma\|_{\infty} \ge \frac{\eta}{2} \sqrt{\dfrac{2\alpha}{3}\dfrac{d^2\log d}{nm^2}}\right] \ge \frac{\sqrt{d}}{1+\sqrt{d}} \left(1 - 2\alpha - \sqrt{\frac{2\alpha}{\log d}}\right).
\end{align*}
}{
\begin{align*}
\inf_{\hat{\Sigma}} \sup_{\Sigma} & \PP_{\Sigma}\left[\|\hat{\Sigma} - \Sigma\|_{\infty} \ge \frac{\eta}{2} \sqrt{\dfrac{2\alpha}{3}\dfrac{d^2\log d}{nm^2}}\right] \\
& \ge \frac{\sqrt{d}}{1+\sqrt{d}} \left(1 - 2\alpha - \sqrt{\frac{2\alpha}{\log d}}\right).
\end{align*}
}
We set $\alpha = 1/10$ and provided that $d \ge 2$, the bound becomes,
\begin{align*}
\inf_{\hat{\Sigma}} \sup_{\Sigma} \PP_{\Sigma}\left[\|\hat{\Sigma} - \Sigma\|_{\infty} \ge \frac{\eta}{2} \sqrt{\dfrac{1}{15}\dfrac{d^2\log d}{nm^2}}\right] \ge \frac{1}{7}.
\end{align*}
Theorem~\ref{thm:infinity_lower} follows now by application of Markov's inequality. 

\medskip
\noindent \textbf{Proof of Theorem~\ref{thm:spectral_lower}:}
As before, the proof is based on an application of Theorem~\ref{thm:tsybakov_multi}, but we will use exponentially many distributions. 
The first distribution $P^n_0$ has the $n$ data vectors drawn iid from $\Ncal(0, \Sigma_0)$ where $\Sigma_0 = \eta I$.
For the remaining distributions, let $\{v_j\}_{j=1}^M$ be the $1/2$-packing in the projection metric guaranteed by Lemma~\ref{lem:packing}. 
We know that $M = \lceil \exp(-1/8)2^{d/8}\rceil$. 
For each $j$, let $P^n_j$ be the distribution where the $n$ data vectors are drawn iid from $\Ncal(0, \Sigma_j)$ where $\Sigma_j = \eta I - \gamma v_jv_j^T$. 

Since we are using a $1/2$-packing, this construction ensures that $\|\Sigma_j - \Sigma_k\|_2 \geq \gamma/2$ for all $j \ne k$, and so the first condition of Theorem~\ref{thm:tsybakov_multi} is satisfied with $s = \gamma/4$. 
All covariance matrices have spectral norm at most $\eta$, and we require $\gamma \le \frac{e-1}{e}\eta$. 
Lastly, by Lemma~\ref{lem:kl_bound}, we have that the average KL is at most $\frac{3}{2}\frac{\gamma^2}{\eta^2}\frac{nm^2}{d^2}$. 
Plugging in for $M$, we require,
\begin{align*}
\frac{3}{2}\frac{\gamma^2}{\eta^2}\frac{nm^2}{d^2} \le \alpha\left(\frac{d}{8}\log 2 - \frac{1}{8}\right),
\end{align*}
which is satisfied if we set $\gamma^2 = \frac{\alpha \eta^2}{48}\frac{d^3}{nm^2}$, provided that $d \ge 4$. 
Setting $\alpha = 1/24$ and provided $d \ge 10$, the right hand side of Theorem~\ref{thm:tsybakov_multi} is lower bounded by $1/3$ while the separation is $\frac{\eta}{4}\sqrt{\frac{1}{1152}\frac{d^3}{nm^2}}$. 
The condition involving $n,m$, and $d$ is based on requiring that $\gamma \le \frac{e-1}{e}\eta$ as required by Lemma~\ref{lem:kl_bound}, and we may apply the Lemma~\ref{lem:packing} since $d \ge 10$. 
The constant $\frac{1}{408}$ is a lower bound on $\frac{1}{3}\times\frac{1}{4}\sqrt{\frac{1}{1152}}$.
Theorem~\ref{thm:spectral_lower} follows by application of Markov's inequality. 

\medskip
\noindent \textbf{Proof of Theorem~\ref{thm:dfree_lb}:}
The goal of this proof is to apply the lower bounds in Theorem~\ref{thm:infinity_lower} and~\ref{thm:spectral_lower} but we will have to show that introducing the distribution and the expectation does not significantly affect that proof.
Recall that in the distribution free setting, the goal is to estimate the sample covariance of the data sequence $x_1,\ldots, x_n$, which could be adversarially generated, but could also be generated according to a particular Gaussian.
In this latter case, the only barrier to applying Theorems~\ref{thm:infinity_lower} and~\ref{thm:spectral_lower} is that we are trying to estimate the sample covariance instead of the population one.
However, since the sample covariance is very close to the population covariance due to concentration of measure, and since our lower bounds guarantee that any estimate is far from the population covariance, it must also be the case that any estimate is far from the sample covariance.
This high-level argument, which we now formalize, gives a lower bound for estimating the sample covariance, which is a distribution free problem. 

To simplify the notation, we will suppress dependence on the projection operators and the sample itself and we write $\hat{\Sigma}$ for the estimator, $\bar{\Sigma}$ for the sample covariance, and $\Sigma$ for the population covariance.
In other words we replace the supremum over samples $x_1^n$ with a supremum over positive semidefinite matrices $\bar{\Sigma}$. 
We must also make sure that $x_1^n$ meets the norm constraints, which for now we will denote by the term ``valid."
Letting $\| \cdot \|$ denote the particular norm of interest, i.e. spectral or $\ell_{\infty}$, we have
\begin{align*}
& \inf_{\hat{\Sigma}} \sup_{\bar{\Sigma}} \PP[\|\hat{\Sigma} - \bar{\Sigma}\| \ge \epsilon_1 - \epsilon_2] \\
& \ge \inf_{\hat{\Sigma}} \sup_{\Sigma} \PP_{\bar{\Sigma} \sim P_{\Sigma}}\left[\|\hat{\Sigma} - \bar{\Sigma}\| \ge \epsilon_1 - \epsilon_2 \bigcap x_1^n\  \textrm{valid}\right]\\
& \ge \inf_{\hat{\Sigma}} \sup_{\Sigma} \PP_{\bar{\Sigma} \sim P_{\Sigma}}[\|\hat{\Sigma} - \bar{\Sigma}\| \ge \epsilon_1 - \epsilon_2] - \PP_{\bar{\Sigma} \sim P_\Sigma}[x_1^n\  \textrm{invalid}].
\end{align*}
In the first line, the only randomness is in the projection operators, while in the second line we have lower bounded the supremum over sample covariances by a supremum over population covariances and an expectation over sample covariances.
The distribution $\bar{\Sigma} \sim P_\Sigma$ is the distribution over $\bar{\Sigma} = \frac{1}{n}\sum_{t=1}^n x_tx_t^T$ where $x_t \overset{i.i.d.}{\sim} \Ncal(0, \Sigma)$.
The second inequality is based on De Morgan's identity and a union bound, and the consequence is we have separated the norm constraints of the sample from the estimation problem altogether.
To bound the probability of the first event, note that by the triangle inequality, if $\|\hat{\Sigma} - \Sigma\| \ge \epsilon_1$ and $\|\Sigma - \bar{\Sigma}\| \le \epsilon_2$, then $\|\hat{\Sigma} - \bar{\Sigma}\| \ge \epsilon_1 - \epsilon_2$.
This means
\begin{align*}
& \PP_{\bar{\Sigma} \sim P_{\Sigma}}[\|\hat{\Sigma} - \bar{\Sigma}\| \ge \epsilon_1 - \epsilon_2]\\
& \ge \PP_{\bar{\Sigma} \sim P_{\Sigma}}[\|\hat{\Sigma} - \Sigma\| \ge \epsilon_1 \bigcap \|\Sigma - \bar{\Sigma}\| \le \epsilon_2]\\
& \ge \PP_{\bar{\Sigma} \sim P_{\Sigma}}[\|\hat{\Sigma} - \Sigma\| \ge \epsilon_1] - \PP_{\bar{\Sigma} \sim P_{\Sigma}}[\|\Sigma - \bar{\Sigma}\| > \epsilon_2].
\end{align*}
The second inequality follows from De Morgan's identity and a union bound. 
Note that the first term is the error in estimating the population covariance given compressed Gaussian samples, so we can apply our lower bounds from before.
The second term does not depend on the estimator or the projections, and it can be controlled by standard concentration-of-measure arguments. 
Putting the terms together gives
\ifthenelse{\equal{\numcolumns}{one}}{
\begin{align*}
& \inf_{\hat{\Sigma}} \sup_{\bar{\Sigma}} \PP[\|\hat{\Sigma} - \bar{\Sigma}\| \ge \epsilon_1 - \epsilon_2] \ge \\
& \inf_{\hat{\Sigma}} \sup_{\Sigma} \PP[\|\hat{\Sigma} - \Sigma\| \ge \epsilon_1] - \PP[\|\Sigma - \bar{\Sigma}\| > \epsilon_2] - \PP[x_1^n\ \textrm{invalid}].
\end{align*}
}{
\begin{align*}
& \inf_{\hat{\Sigma}} \sup_{\bar{\Sigma}} \PP[\|\hat{\Sigma} - \bar{\Sigma}\| \ge \epsilon_1 - \epsilon_2] \ge \\
& \inf_{\hat{\Sigma}} \sup_{\Sigma} \PP[\|\hat{\Sigma} - \Sigma\| \ge \epsilon_1] - \PP[\|\Sigma - \bar{\Sigma}\| > \epsilon_2]\\
 & \qquad - \PP[x_1^n\ \textrm{invalid}].
\end{align*}
}
We now derive the $\ell_{\infty}$ norm bound.
By examining the proof of Theorem~\ref{thm:infinity_upper} we know that with $\epsilon_1 = \frac{\eta}{2}\sqrt{\frac{1}{15}\frac{d^2\log(d)}{nm^2}}$ the first term can be lower bounded by $\frac{1}{7}$ provided that $\Sigma$ is allowed to have $\ell_{\infty}$ norm as large as $\eta$. 
For the second term, we can apply the fourth inequality in Proposition~\ref{prop:gaussian_tails} and set $\epsilon_2 = \eta\sqrt{\frac{\log(2d/\delta)}{n}}$ so that this probability is at most $\delta$. 
Finally the third term constrains our setting of $\eta$, the infinity norm bound of the population covariance.
The sample is valid if $\|X\|_{\infty}^2 \le \eta_{\infty}$ and the first inequality in Proposition~\ref{prop:gaussian_tails} shows that if
\begin{align*}
\eta \le \frac{\eta_\infty}{2 \log(nd/\delta)}
\end{align*}
then the sample will be invalid with probability at most $\delta$.
Setting $\delta = \frac{1}{28}$ and $\eta$ to meet the inequality reveals that with probability at least $\frac{1}{14}$, there is a constant $\kappa_1 > 0$ such that
\begin{align*}
\|\hat{\Sigma} - \bar{\Sigma}\|_{\infty} \ge \frac{\kappa_1\eta_\infty}{\log(nd)}\left(\sqrt{\frac{d^2\log(d)}{nm^2}} - \sqrt{\frac{\log(d)}{n}}\right).
\end{align*}
This bound holds in a minimax sense and the first claim in Theorem~\ref{thm:dfree_lb} follows from Markov's inequality.
This bound requires the conditions on $n$ and $d$ from Theorem~\ref{thm:infinity_lower}.

The two spectral norm bounds follow in a similar manner.
We can set $\epsilon_1 = \frac{\eta}{408}\sqrt{\frac{d^3}{nm^2}}$ so that the first term is lower bounded by $\frac{1}{3}$, provided that $\Sigma$ is allowed to have spectral norm as large as $\eta$ in the construction. 
The second term can be bounded by the third inequality in Proposition~\ref{prop:gaussian_tails}: with $\epsilon_2 = \eta \sqrt{\frac{c d\log(2/\delta)}{n}}$ the probability is at most $\delta$.

For Equation~\eqref{eq:dfree_spectral_1}, the sample is valid if $\|\bar{\Sigma}\|_2\|X\|_{2,\infty}^2 \le \eta_{S_1}$
By applying the second inequality of Proposition~\ref{prop:gaussian_tails} the left hand side is bounded by
\begin{align*}
\ifthenelse{\equal{\numcolumns}{one}}{
 \|\bar{\Sigma}\|_2 \|X\|_{2,\infty}^2 \le 2d \|\Sigma\|_2 \log(nd/\delta) \left(\|\bar{\Sigma} - \Sigma\|_2 + \|\Sigma\|_2\right)
 \le 2d\|\Sigma\|_2^2\log(nd/\delta)\left(1 + \sqrt{\frac{cd\log(2/\delta)}{n}}\right).
}{
 \|\bar{\Sigma}\|_2 \|X\|_{2,\infty}^2 & \le 2d \|\Sigma\|_2 \log(nd/\delta) \left(\|\bar{\Sigma} - \Sigma\|_2 + \|\Sigma\|_2\right)\\
 & \le 2d\|\Sigma\|_2^2\log(nd/\delta)\left(1 + \sqrt{\frac{cd\log(2/\delta)}{n}}\right).
}
\end{align*}
Setting $\delta = 1/12$ and if $n \ge d\log(2/\delta)$, we may set
\begin{align*}
\eta = c \sqrt{\frac{\eta_{S_1}}{d\log(nd)}}.
\end{align*}
This choice implies that there is some constant $\kappa_2$ such that, with probability at least $1/6$, we have
\begin{align*}
\|\hat{\Sigma} - \bar{\Sigma}\|_2 &\ge \kappa_2 \sqrt{\frac{\eta_{S_1}}{d \log(nd)}} \left(\sqrt{\frac{d^3}{nm^2}} - \sqrt{\frac{d}{n}}\right).
\end{align*}
Finally, we use the fact that $d^2/m^2 \ge d/m$ since $m \le d$. 

For Equation~\eqref{eq:dfree_spectral_2}, the sample is valid if $\frac{1}{n}\sum_{t=1}^n\|x_t\|_2^4 \le \eta_{S_2}$.
Again, the second inequality in Proposition~\ref{prop:gaussian_tails} gives
\begin{align*}
\ifthenelse{\equal{\numcolumns}{one}}{
\frac{1}{n}\sum_{t=1}^n \|x_t\|_2^4 \le \max_{t \in [n]} \|x_t\|_2^4 \le (2 \tr(\Sigma) \log(nd/\delta))^2  \le 4 d^2 \|\Sigma\|^2_2 \log^2(nd/\delta).
}{
\frac{1}{n}\sum_{t=1}^n \|x_t\|_2^4 & \le \max_{t \in [n]} \|x_t\|_2^4 \le (2 \tr(\Sigma) \log(nd/\delta))^2  \\
& \le 4 d^2 \|\Sigma\|^2_2 \log^2(nd/\delta).
}
\end{align*}
Setting $\delta = 1/12$, we want this to be at most $\eta_{S_2}$ which means we may set $\eta$ to be
\begin{align*}
\eta = c \frac{\sqrt{\eta_{S_2}}}{d\log(nd)}.
\end{align*}
As in the previous case, this implies the lower bound
\begin{align*}
\|\hat{\Sigma} - \bar{\Sigma}\|_2 &\ge \kappa_3\frac{\sqrt{\eta_{S_2}}}{d\log(nd)} \left(\sqrt{\frac{d^3}{nm^2}} - \sqrt{\frac{d}{n}}\right).
\end{align*}
This holds for some $\kappa_3 > 0$ with probability at least $\frac{1}{6}$. 

\medskip
\noindent \textbf{Proof of Proposition~\ref{prop:fixed}:}
To prove Proposition~\ref{prop:fixed}, we need one intermediate result.
The following lemma lower bounds the minimax risk (in squared $\ell_2$ norm) of estimating a vector when it is observed via a low-dimensional random projection.
\begin{lemma}[Lower bound for compressed vector estimation]
\label{lem:compressed_vector_estimation}
Let $\Pcal(\nu)$ be the uniform distribution over vectors with $\ell_2$ norm $\nu$ in $\RR^d$ and let $\Ucal_m$ be the uniform distribution over $m$-dimensional projection matrices over $\RR^d$.
Then
\begin{align*}
\ifthenelse{\equal{\numcolumns}{one}}{
\inf_{T} \sup_{x: \|x\|_2 = \nu} \EE_{\Pi \sim \Ucal_m} \|T(\Pi, \Pi x) - x\|_2^2 \ge
\inf_{T} \EE_{x \sim \Pcal(\nu)} \EE_{\Pi \sim \Ucal_m} \| T(\Pi, \Pi x) - x\|_2^2 \ge \nu^2 \left(1 - \frac{m}{d}\right).
}{
& \inf_{T} \sup_{x: \|x\|_2 = \nu} \EE_{\Pi \sim \Ucal_m} \|T(\Pi, \Pi x) - x\|_2^2 \ge\\
& \inf_{T} \EE_{x \sim \Pcal(\nu)} \EE_{\Pi \sim \Ucal_m} \| T(\Pi, \Pi x) - x\|_2^2 \ge \nu^2 \left(1 - \frac{m}{d}\right).
}
\end{align*}
\end{lemma}

For the theorem, we lower bound the minimax risk by
\ifthenelse{\equal{\numcolumns}{one}}{
\begin{align*}
\inf_{T} \sup_{\Sigma, \|\Sigma\|_2 \le \eta} \EE_{\Pi \sim \Ucal_m} \|T(\Pi, \Pi \Sigma \Pi) - \Sigma\|_2 & \ge \inf_{T} \EE_{x \sim \Pcal(\sqrt{\eta})} \EE_{\Pi \sim \Ucal_m} \|T(\Pi, \Pi xx^T \Pi) - xx^T\|_2\\
& \ge \eta \times \inf_{T} \EE_{\stackrel{x \sim \Pcal(\sqrt{\eta})}{\Pi \sim \Ucal_m}} \sin \angle \left(v_{\max}(T(\Pi, \Pi xx^T \Pi)), \frac{x}{\|x\|_2}\right),
\end{align*}
}{
\begin{align*}
& \inf_{T} \sup_{\Sigma, \|\Sigma\|_2 \le \eta} \EE_{\Pi \sim \Ucal_m} \|T(\Pi, \Pi \Sigma \Pi) - \Sigma\|_2 \\
& \ge \inf_{T} \EE_{x \sim \Pcal(\sqrt{\eta})} \EE_{\Pi \sim \Ucal_m} \|T(\Pi, \Pi xx^T \Pi) - xx^T\|_2\\
& \ge \eta \times \inf_{T} \EE_{\stackrel{x \sim \Pcal(\sqrt{\eta})}{\Pi \sim \Ucal_m}} \sin \angle \left(v_{\max}(T(\Pi, \Pi xx^T \Pi)), \frac{x}{\|x\|_2}\right),
\end{align*}
}
where $v_{\max}(M)$ is the eigenvector corresponding to the largest eigenvalue of $M$. 
Here we are applying Theorem~\ref{thm:davis_kahan} and using the eigengap for the matrix $xx^T$, which is $\eta$. 
This bound holds for any leading eigenvector of $T(\Pi,\Pi xx^T\Pi)$, and we apply it with the eigenvector that has non-negative inner product with $x$. 

For two unit vectors $v$ and $y$ with non-negative inner product, we have that
\begin{align*}
  \sin^2 \angle (v,y) = 1 - (v^Ty)^2 \ge 1 - v^Ty = \|v-y\|_2^2/2.
\end{align*}
Since $v_{\max}$ is the leading eigenvector with non-negative inner product with $x$, it is unit-normed, so we are able to translate to the Euclidean norm and obtain
\ifthenelse{\equal{\numcolumns}{one}}{
\begin{align*}
\eta \times \inf_{T}\EE_{\stackrel{x \sim \Pcal(\sqrt{\eta}))}{\Pi \sim \Ucal_m}} \frac{1}{\sqrt{2}}\left\| v_{\max}(T(\Pi,\Pi xx^T\Pi)) - \frac{x}{\|x\|_2} \right\|_2
& \ge \frac{\eta}{\sqrt{2}} \times \inf_{v: \|v\|=1} \EE_{\stackrel{x \sim \Pcal(1)}{\Pi \sim \Ucal_m}} \|v(\Pi,\Pi x) - x\|_2
\end{align*}
}{
\begin{align*}
& \eta \times \inf_{T}\EE_{\stackrel{x \sim \Pcal(\sqrt{\eta}))}{\Pi \sim \Ucal_m}} \frac{1}{\sqrt{2}}\left\| v_{\max}(T(\Pi,\Pi xx^T\Pi)) - \frac{x}{\|x\|_2} \right\|_2\\
& \ge \frac{\eta}{\sqrt{2}} \times \inf_{v: \|v\|=1} \EE_{\stackrel{x \sim \Pcal(1)}{\Pi \sim \Ucal_m}} \|v(\Pi,\Pi x) - x\|_2
\end{align*}
}
Here we instead take infimum over estimators $v(\Pi, \Pi x)$ for the vector $x$, and we draw $x$ uniformly from the unit sphere.
This lower bounds the error because the leading eigenvector of $\Pi xx^T\Pi$ is (up to sign) $\Pi x$, so we are only providing additional information to the estimator $v$. 
Proposition~\ref{prop:fixed} now follows by applying Lemma~\ref{lem:compressed_vector_estimation}.
Notice that we apply Lemma~\ref{lem:compressed_vector_estimation} with $\nu = 1$ since the application of the Davis-Kahan theorem already accounts for normalization.

\section{Conclusions}
\label{sec:conclusion}
In this paper, we studied the problem of estimating a covariance matrix from highly compressive measurements. 
We proposed an estimator based on projecting the observations back into the high-dimensional space, and we bounded the infinity and spectral norm error of this estimator.
We complemented this analysis with minimax lower bounds for this problem, showing that our estimator is rate-optimal.
We showed that this estimator also adapts to low rank structure in the target covariance, and we mentioned applications to subspace learning and to learning in distributed sensor networks.
Note that many other consequences are immediate, for example the task of learning the structure of a Gaussian graphical model follows from the results of Ravikumar et al.~\cite{ravikumar2011high}. 

The main insight of our work is that by leveraging independent random projection operators for each data point, we can build consistent covariance estimators from compressive measurements even in an unstructured setting. 
However, due to the absence of structure in this problem, the effective sample size shrinks from $n$ in the classical setting to $nm^2/d^2$.
This gives a precise characterization of the effects of data compression in the covariance estimation problem.

\section*{Acknowledgements}
This work is supported by NSF award IIS-1247658 and CAREER IIS-1252412 and an AFOSR YIP award.

\bibliography{mc}

\begin{thebibliography}{10}

\bibitem{achlioptas2007fast}
Dimitris Achlioptas and Frank Mcsherry.
\newblock {Fast computation of low-rank matrix approximations}.
\newblock {\em Journal of the ACM}, April 2007.

\bibitem{anantharam2013maximal}
Venkat Anantharam, Amin Gohari, Sudeep Kamath, and Chandra Nair.
\newblock {On maximal correlation, hypercontractivity, and the data processing
  inequality studied by Erkip and Cover}.
\newblock {\em arXiv:1304.6133}, 2013.

\bibitem{cai2015rop}
T.~Tony Cai and Anru Zhang.
\newblock {ROP: Matrix recovery via rank-one projections}.
\newblock {\em The Annals of Statistics}, 2015.

\bibitem{candes2006robust}
Emmanuel~J. Cand\`{e}s, Justin Romberg, and Terence Tao.
\newblock {Robust uncertainty principles: Exact signal reconstruction from
  highly incomplete frequency information}.
\newblock {\em IEEE Transactions on Information Theory}, 2006.

\bibitem{chen2013exact}
Yuxin Chen, Yuejie Chi, and Andrea Goldsmith.
\newblock Exact and stable covariance estimation from quadratic sampling via
  convex programming.
\newblock {\em {IEEE Transactions on Information Theory}}, 2015.

\bibitem{dasarathy2013sketching}
Gautam Dasarathy, Parikshit Shah, Badri~Narayan Bhaskar, and Robert~D. Nowak.
\newblock Sketching sparse matrices, covariances, and graphs via tensor
  products.
\newblock {\em IEEE Transactions on Information Theory}, 2015.

\bibitem{dasgupta2003elementary}
Sanjoy Dasgupta and Anupam Gupta.
\newblock {An elementary proof of a theorem of Johnson and Lindenstrauss}.
\newblock {\em Random Structures \& Algorithms}, 2003.

\bibitem{davis1970rotation}
Chandler Davis and W.~M. Kahan.
\newblock The rotation of eigenvectors by a perturbation. {III}.
\newblock {\em SIAM Journal on Numerical Analysis}, 1970.

\bibitem{delapena2012decoupling}
Victor {De la Pe\~{n}a} and Evarist Gin\'{e}.
\newblock {\em {Decoupling: from dependence to independence}}.
\newblock Springer Science \& Business Media, 2012.

\bibitem{donoho2006compressed}
David~L. Donoho.
\newblock {Compressed sensing}.
\newblock {\em IEEE Transactions on Information Theory}, 2006.

\bibitem{duchi2013local}
John~C. Duchi, Michael~I. Jordan, and Martin~J. Wainwright.
\newblock Local privacy and statistical minimax rates.
\newblock In {\em Foundations of Computer Science}, 2013.

\bibitem{ericson2001codes}
Thomas Ericson and Victor Zinoviev.
\newblock {\em {Codes on Euclidean spheres}}.
\newblock Elsevier, 2001.

\bibitem{gonen2014sample}
Alon Gonen, Dan Rosenbaum, Yonina Eldar, and Shai Shalev-Shwartz.
\newblock Subspace learning with partial information.
\newblock {\em Journal of Machine Learning Research}, 2016.

\bibitem{halko2011finding}
Nathan Halko, Per-Gunnar Martinsson, and Joel~A. Tropp.
\newblock {Finding structure with randomness: Probabilistic algorithms for
  constructing approximate matrix decompositions}.
\newblock {\em SIAM Review}, 2011.

\bibitem{kolar2012consistent}
Mladen Kolar and Eric~P. Xing.
\newblock {Consistent covariance selection from data with missing values}.
\newblock In {\em International Conference on Machine Learning}, 2012.

\bibitem{krishnamurthy2014subspace}
Akshay Krishnamurthy, Martin Azizyan, and Aarti Singh.
\newblock {Subspace learning from extremely compressed measurements}.
\newblock In {\em Asilomar Conference on Signals Systems and Computers}, 2014.

\bibitem{liberty2007randomized}
Edo Liberty, Franco Woolfe, Per-Gunnar Martinsson, Vladimir Rokhlin, and Mark
  Tygert.
\newblock {Randomized algorithms for the low-rank approximation of matrices.}
\newblock {\em Proceedings of the National Academy of Sciences}, 2007.

\bibitem{loh2012high}
Po-Ling Loh and Martin~J. Wainwright.
\newblock High-dimensional regression with noisy and missing data: Provable
  guarantees with nonconvexity.
\newblock {\em {The Annals of Statistics}}, 2012.

\bibitem{maurer2003bound}
Andreas Maurer.
\newblock {A bound on the deviation probability for sums of non-negative random
  variables}.
\newblock {\em Journal of Inequalities in Pure and Applied Mathematics}, 2003.

\bibitem{negahban2012restricted}
Sahand Negahban and Martin~J. Wainwright.
\newblock {Restricted strong convexity and weighted matrix completion: optimal
  bounds with noise}.
\newblock {\em The Journal of Machine Learning Research}, 2012.

\bibitem{pourkamali2014memory}
Farhad Pourkamali-Anaraki and Shannon Hughes.
\newblock {Memory and computation efficient PCA via very sparse random
  projections}.
\newblock In {\em International Conference on Machine Learning}, 2014.

\bibitem{raginsky2014strong}
Maxim Raginsky.
\newblock Strong data processing inequalities and $\phi$-sobolev inequalities
  for discrete channels.
\newblock {\em IEEE Transactions on Information Theory}, 2014.

\bibitem{ravikumar2011high}
Pradeep Ravikumar, Martin~J. Wainwright, Garvesh Raskutti, and Bin Yu.
\newblock {High-dimensional covariance estimation by minimizing
  $\ell_1$-penalized log-determinant divergence}.
\newblock {\em Electronic Journal of Statistics}, 2011.

\bibitem{recht2011simpler}
Benjamin Recht.
\newblock {A simpler approach to matrix completion}.
\newblock {\em The Journal of Machine Learning Research}, 2011.

\bibitem{sarlos2006improved}
Tamas Sarlos.
\newblock {Improved approximation algorithms for large matrices via random
  projections}.
\newblock In {\em Foundations of Computer Science}, 2006.

\bibitem{tropp2011user}
Joel~A. Tropp.
\newblock {User-friendly tail bounds for sums of random matrices}.
\newblock {\em Foundations of Computational Mathematics}, August 2011.

\bibitem{tsybakov2009introduction}
Alexandre~B. Tsybakov.
\newblock {\em {Introduction to nonparametric estimation}}.
\newblock Springer, 2009.

\bibitem{uspensky1937introduction}
James~Victor Uspensky.
\newblock {\em {Introduction to mathematical probability}}.
\newblock McGraw-Hill, 1937.

\bibitem{vershynin2010introduction}
Roman Vershynin.
\newblock Introduction to the non-asymptotic analysis of random matrices.
\newblock {\em Compressed Sensing}, 2012.

\end{thebibliography}
\bibliographystyle{plain}

\vfill
\newpage 
\appendix
\section{Deviation Bounds}
\label{sec:app_dev}

\begin{theorem}[Bernstein Inequality]
\label{thm:bernstein}
If $U_1, \ldots, U_n$ are independent zero-mean random variables with $|U_t| \le B$ a.s. and $\dfrac{1}{n}\sum_{t=1}^n\Var(U_t) \le \sigma^2$, then, for any $\delta \in (0,1)$,
\begin{align*}
\PP\left(\left|\frac{1}{n}\sum_{t=1}^nU_t\right| \le \sqrt{\frac{2\sigma^2\log(2/\delta)}{n}} + \frac{2B\log(2/\delta)}{3n}\right) \ge 1-\delta.
\end{align*}
\end{theorem}

\begin{theorem}[Matrix Bernstein Inequality~\cite{tropp2011user}]
\label{thm:matrix_bernstein}
Let $X_1, \ldots, X_n$ be independent, random, self-adjoint matrices with dimension $d$ satisfying
\[
\EE X_k = 0 \qquad \textrm{and} \qquad \|X_k\|_2 \le R \textrm{ almost surely}.
\]
Then, for all $t \ge 0$,
\[
\PP\left(\left\|\sum_{k=1}^nX_k\right\| \ge t\right) \le d \exp\left( \frac{-t^2/2}{\sigma^2 + Rt/3}\right), \qquad \textrm{where} \qquad \sigma^2= \left\|\sum_{k=1}^n \EE X_k^2\right\|.
\]
\end{theorem}

\begin{theorem}[Subexponential Matrix Bernstein Inequality~\cite{tropp2011user}]
\label{thm:matrix_bernstein_subexp}
Let $X_1, \ldots, X_n$ be independent, random, mean zero, symmetric matrices with dimension $d$.
Assume that there exists $R \in \RR$ and $A_k \in \RR^{d \times d}$, $k \in [n]$ such that
\begin{align*}
\EE(X_k^p) \preceq \frac{p!}{2} R^{p-2} A_k^2 \qquad \textrm{for } p = 2, 3, 4, \ldots .
\end{align*}
Then, for all $t \ge 0$,
\begin{align*}
\PP\left(\left\|\sum_{k=1}^nX_k\right\| \ge t\right) \le d \exp\left( \frac{-t^2/2}{\sigma^2 + Rt}\right), \qquad \textrm{where} \qquad \sigma^2= \left\|\sum_{k=1}^n A_k^2\right\|.
\end{align*}
\end{theorem}

\begin{theorem}[\cite{maurer2003bound}]
\label{thm:positive_tail}
Let $X_1, \ldots, X_n$ be independent random variables with $\EE X_t^2 < \infty$ and $X_t \ge 0$, a.s., for all $t \in [n]$.
Then, for any $\epsilon > 0$,
\begin{align*}
\PP\left( \EE\left(\sum_{t=1}^nX_t\right) - \sum_{t=1}^n X_t \ge \epsilon\right) \le \exp\left(\frac{-\epsilon^2}{2\sum_{t=1}^n\EE X_t^2}\right).
\end{align*}
\end{theorem}

\begin{theorem}[\cite{uspensky1937introduction}]
\label{thm:mgf_bound}
Let $X_1, \ldots, X_n$ be independent random variables with $\EE X_t = 0$ and $\EE X_t^2 = b_t$ and
\begin{align*}
\EE|X_t|^p \le \frac{p!}{2}a^{p-2}b_t,
\end{align*}
for all $t\in [n]$, $p \ge 3$ and for some constant $a > 0$. 
Then
\begin{align*}
  \EE \exp\left(\lambda \sum_{t=1}^nX_t\right) \le \exp\left(\frac{\lambda^2 \sum_{t=1}^nb_t}{2(1-s)}\right),
\end{align*}
for any $s \in (0,1)$ and $\lambda > 0$ provided that $\lambda a \le s$. 
\end{theorem}

\begin{proposition}
\label{prop:beta_dev_1}
For any $t \in [n]$ and $d \ge 3$, let $c_t \ge 0$ and $\nu_t \sim \textrm{Beta}(\frac{1}{2}, \frac{d-2}{2})$.
Define $c \triangleq \max_{t \in [n]} c_t$ and
\begin{align*}
B \triangleq \frac{2(d-2)nc^4}{(d-1)^2(d+1)}.
\end{align*}
For any $s \in (0,1)$ and $\delta > 0$, if $\log(1/\delta) \le \frac{(d+3)^2Bs^2}{3200c^4(1-s)}$, then
\begin{align*}
\PP\left(\frac{1}{n}\sum_{t=1}^n c_t^2\nu_t - \frac{1}{n}\sum_{t=1}^n\frac{1}{d-1}c_t^2 \ge \sqrt{\frac{2B\log(1/\delta)}{n^2(1-s)}} \right)\le \delta.
\end{align*}
\end{proposition}
\ifthenelse{\equal{\version}{arxiv}}{
\begin{proof}
}{
\begin{IEEEproof}
}
For $t \in [n]$, define $X_t = c_t^2\left(\nu_t - \frac{1}{d-1}\right)$ and note that $\EE X_t = 0$. 
Let $b_t = \EE X_t^2 = c_t^4\dfrac{2(d-2)}{(d-1)^2(d+1)}$. 
Then for any $k \ge 3$ we have, using Minkowski's Inequality,
\begin{align*}
\EE |X_t|^k &= c_t^{2k}\EE\left|\nu_t - \frac{1}{d-1}\right|^{k}
\le c_t^{2k}\left( (\EE\nu_t^k)^{1/k} + \frac{1}{d-1}\right)^{k}\\
& = c_t^{2k}\left(\left(\prod_{r=0}^{k-1} \frac{1/2 + r}{(d-1)/2+r} \right)^{1/k} + \frac{1}{d-1}\right)^k.
\end{align*}
Here we also use the fact that the $k$th moment of $\textrm{Beta}(\alpha,\beta)$ random variable is $\prod_{r=0}^{k-1} \frac{\alpha+r}{\alpha+\beta+r}$. 
We now leverage two claims.
For what remains of the proof, we use the notation $\prod_{r=3}^2\left(\cdot\right) = 1$.

\textbf{Claim 1.}
For $k \ge 3$ and $d \ge 3$,
\begin{align*}
\max\left(\frac{1/2+k}{(d-1)/2+k} \times \frac{1}{k+1}, \frac{d-1}{d-2}\frac{10}{(d+3)}\right) \le \frac{20}{d+3}.
\end{align*}
{\em Proof of claim.}
Since $d \ge 3$ the second term is clearly at most $\frac{20}{d+3}$.
For the first term, after rearranging, we require
\begin{align*}
(1+2k) (d+3) \le 20(k+1)(d-1+2k)
& \Leftrightarrow 0 \le 40k^2 + 14k + 18kd + 19d - 23
\end{align*}
The expression on the right hand side is convex in $k$ and, by taking derivative, it is minimized when $k = \frac{-18d-14}{80}$ and hence it is monotonically increasing for $k \ge 3$. 
Clearly it is also monotonically increasing in $d$ as well, and plugging in $k=d=3$ verifies the inequality.

\hfill\ensuremath{\blacksquare}

\textbf{Claim 2.}
For any $\zeta \ge \frac{20}{d+3}$ and $k \ge 3$,
\begin{align*}
\frac{d-1}{d-2}\frac{60}{d+3}\prod_{r=3}^{k-1}\frac{1/2+r}{(d-1)/2+r} \le k!\zeta^{k-2}.
\end{align*}
{\em Proof of claim.}
We proceed by induction.
For $k=3$, the expression simplifies to the second term in the previous claim,
\begin{align*}
\frac{d-1}{d-2}\frac{60}{d+3} \le 6 \zeta.
\end{align*}
For the inductive step, assume the claim holds for $k \ge 3$. Then for $k+1$, we have
\begin{align*}
\frac{d-1}{d-2}\frac{60}{d+3}\prod_{r=3}^{k}\frac{1/2+r}{(d-1)/2+r} \le \frac{1/2+k}{(d-1)/2+k}k!\zeta^{k-2} \le (k+1)! \zeta^{k-1},
\end{align*}
where the first step is the inductive hypothesis and the second is from the first part of the previous claim.

\hfill\ensuremath{\blacksquare}

Equipped with these two claims we can proceed with the proof of the proposition.
For any $a_t \ge \frac{40c_t^2}{d+3}$, we have
\begin{align*}
\EE |X_t|^k & \le c_t^{2k}\left(\left(\prod_{r=0}^{k-1} \frac{1/2+r}{(d-1)/2+r}\right)^{1/k} + \frac{1}{d-1}\right)^{k}\\
& \le 2^kc_t^{2k}\prod_{r=0}^{k-1} \frac{1/2+r}{(d-1)/2+r} = 2^kc_t^{2k}\frac{15}{(d-1)(d+1)(d+3)} \prod_{r=3}^{k-1} \frac{1/2+r}{(d-1)/2+r}\\
& = \frac{c_t^4}{(d-1)(d+1)} (2c_t^2)^{k-2} \frac{60}{d+3}\prod_{r=3}^{k-1} \frac{1/2+r}{(d-1)/2+r}\\
& \le \frac{c_t^4}{(d-1)(d+1)} (2c_t^2)^{k-2}\frac{d-2}{d-1}k!\left(\frac{20}{d+3}\right)^{k-2}\\
& \le \frac{c_t^4}{(d-1)(d+1)} \frac{d-2}{d-1}k!a_t^{k-2} = \frac{b_t}{2}k!a_t^{k-2}.
\end{align*}
Here the first inequality is from the application of Minkowski's inequality above, the second uses the fact that $\dfrac{1}{d-1} \le \frac{1/2+r}{(d-1)/2+r}$ for any $r$.
In the third line we pull out terms from the product.
Then we apply the claim from before with $\zeta = \dfrac{20}{d+3}$ and finally substitute in for $a_t$ and $b_t$. 

Now setting $a \triangleq \frac{40c^2}{d+3}$, we have that the moment bound above holds for all $X_t$. 
Let $s \in (0,1)$ and $\lambda > 0$ such that $\lambda a \le s$.
Since $B \triangleq \frac{2(d-2)nc^4}{(d-1)^2(d+1)} \ge \sum_{t=1}^n b_t$, by Theorem~\ref{thm:mgf_bound} we have
\begin{align*}
\EE\exp\left(\lambda \sum_{t=1}^nX_t\right) \le \exp\left( \frac{B\lambda^2}{2(1-s)}\right).
\end{align*}
We may now apply the Chernoff trick, so that, for any $\epsilon > 0$,
\begin{align*}
\PP\left(\sum_{t=1}^nX_t \ge \epsilon \right) \le \exp\left( - \lambda \epsilon + \frac{B\lambda^2}{2(1-s)}\right) = \exp\left(\lambda\left(\frac{B\lambda}{2(1-s)} - \epsilon\right)\right).
\end{align*}
Set $\lambda \triangleq \dfrac{1-s}{B}\epsilon$, so that if $\dfrac{1-s}{B}\epsilon \le \dfrac{s}{a}$, we have
\begin{align*}
\PP\left(\sum_{t=1}^nX_t \ge \epsilon \right) \le \exp\left( \frac{-\epsilon^2(1-s)}{2B}\right).
\end{align*}
Inverting this inequality and the condition above proves the result. 
In particular, we require that for the $\delta > 0$ that we choose, $\dfrac{1-s}{B}\sqrt{\dfrac{2B\log(1/\delta)}{1-s}} \le \dfrac{s}{a}$.
If this is the case, we have
\begin{align*}
\PP\left(\sum_{t=1}^n X_t \ge \sqrt{\frac{2B\log(1/\delta)}{1-s}}\right) \le \delta.
\end{align*}
\ifthenelse{\equal{\version}{arxiv}}{
\end{proof}
}{
\end{IEEEproof}
}

\begin{proposition}
\label{prop:beta_dev_2}
For any $t \in [n]$ and $d \ge 3$, let $c_t \ge 0$ and $\nu_t \sim \textrm{Beta}(\frac{1}{2}, \frac{d-2}{2})$.
Define $c = \max_{t \in [n]} c_t$.
Then, for any $\delta > 0$,
\begin{align*}
\PP\left(\frac{1}{n}\sum_{t=1}^n \frac{1}{d-1} c_t^2 - \frac{1}{n}\sum_{t=1}^nc_t^2\nu_t \ge \sqrt{\frac{6c^4\log(1/\delta)}{n(d^2-1)}}\right)\le \delta.
\end{align*}
\end{proposition}
\ifthenelse{\equal{\version}{arxiv}}{
\begin{proof}
}{
\begin{IEEEproof}
}
For each $t \in [n]$, we have $c_t^2\nu_t \ge 0$, $\EE c_t^2\nu_t = \frac{1}{d-1}c_t^2$, and $\EE(c_t^2\nu_t)^2 = c_t^4\frac{3}{d^2-1} \le \frac{3c^4}{d^2-1}$.
So by Theorem~\ref{thm:positive_tail}, for any $\epsilon > 0$, we have
\begin{align*}
\PP\left(\frac{1}{n}\sum_{t=1}^n\frac{1}{d-1}c_t^2 - \frac{1}{n}\sum_{t=1}^nc_t^2\nu_t \ge \frac{\epsilon}{n}\right) \le \exp\left(\frac{-\epsilon^2}{2n\left(\frac{3c^4}{d^2-1}\right)}\right),
\end{align*}
and the result follows by inverting the inequality.
\ifthenelse{\equal{\version}{arxiv}}{
\end{proof}
}{
\end{IEEEproof}
}

\begin{proposition}
\label{prop:beta_dev_3}
If $\nu \sim \textrm{Beta}(\frac{1}{2}, \frac{d-2}{2})$, for $d \ge 4$ then $\PP(\nu > \frac{2}{d-3}\log(1/\delta)) \le \delta$ for any $\delta > 0$. 
\end{proposition}
\ifthenelse{\equal{\version}{arxiv}}{
\begin{proof}
}{
\begin{IEEEproof}
}
Let $\alpha \in \RR^{d-1}$ be uniformly distributed on the unit sphere and let $\zeta \sim \textrm{Beta}(1, \frac{d-3}{2})$. 
Then $\alpha(1)^2 \,{\buildrel d\over{=}}\, \nu$ and $\alpha(1)^2 + \alpha(2)^2 \,{\buildrel d\over{=}}\, \zeta$. 
So, for any $\epsilon \in (0,1)$,
\begin{align*}
\PP(\nu \le \epsilon) = \PP(\alpha(1)^2 \le \epsilon) \ge \PP(\alpha(1)^2+ \alpha(2)^2 \le \epsilon) = \PP(\zeta \le \epsilon) = \PP(-\log(1-\zeta) \le -\log(1-\epsilon)).
\end{align*}
It is well known that $-\log(1-\zeta)$ is exponentially distributed with rate $\frac{d-3}{2}$, and so
\begin{align*}
\PP(-\log(1-\zeta) \le -\log(1-\epsilon)) \ge \PP(-\log(1-\zeta) \le \epsilon) = 1 - \exp\left(-\epsilon \frac{d-3}{2} \right).
\end{align*}
\ifthenelse{\equal{\version}{arxiv}}{
\end{proof}
}{
\end{IEEEproof}
}

\section{Proofs of Technical Lemmas}

\subsection{Detailed proof of Lemma~\ref{lem:quadratic_form}}
Here we provide a detailed proof of the quadratic-form deviation bound.
Recall the claim to be proved: for $d \ge 2$, for any unit vector $u \in \RR^d$ and for any $\delta \in (0,1)$ with $\delta \le n/e$ and $\log(1/\delta) \le \frac{n}{9600} \frac{d(d+3)^2}{(d-1)^2(d+1)}$, with probability $\ge 1-4\delta$ we have
\begin{align*}
\left|u^T(\hat{\Sigma}_1 - \bar{\Sigma})u\right| & \le \sqrt{\frac{d^2 \log(2/\delta)}{nm^2}} \frac{8c^2}{d}
+ \sqrt{\frac{d^2 \log(2/\delta)}{nm^2}} \left[ \sqrt{18 \log^2(n/\delta)} \left(b^2 + \frac{16c^2}{d}\right)\right]\\
+ & \frac{4d^2\log(2/\delta)}{nm^2} \left(b^2 + \frac{c^2\log(n/\delta)}{d}\right),
\end{align*}
where $b \triangleq \max_{t \in [n]} |x_t^Tu|$ and $c \triangleq \max_{t \in [n]} \sqrt{\|x_t\|_2^2 - (x_t^Tu)^2}$. 

Let $b_t \triangleq |x_t^Tu|$ and let $c_t \triangleq \sqrt{\|x_t\|^2 - (x_t^Tu)^2}$. 
Then
\begin{align*}
u^T\bar{\Sigma}u &= \frac{d(dm+d-2)}{m(d+2)(d-1)}u^T\Sigma u + \frac{d (d-m)}{m(d+2)(d-1)}\tr(\Sigma)\\
& = \frac{1}{n}\sum_{t=1}^n\frac{d(m+2)}{m(d+2)}(x_t^Tu)^2 + \frac{d(d-m)}{m(d+2)(d-1)}(\|x_t\|_2^2 - (x_t^Tu)^2)\\
& = \frac{1}{n}\sum_{t=1}^n\frac{d(m+2)}{m(d+2)}b_t^2 + \frac{d(d-m)}{m(d+2)(d-1)}c_t^2.
\end{align*}

We now expand the term involving $\hat{\Sigma}_1$. 
Write $\Phi_t u = \omega_t u + \sqrt{\omega_t - \omega_t^2}W\alpha_t$ where $\omega_t \sim \textrm{Beta}(\frac{m}{2}, \frac{d-m}{2}), \alpha_t \in \RR^{d-1}$ is distributed uniformly on the unit sphere and independently from $\omega_t$, and $W$ is an orthonormal basis for the subspace orthogonal to $u$. 
Then
\begin{align*}
u^T\hat{\Sigma}_1u &\,{\buildrel d \over =}\, \frac{d^2}{nm^2}\sum_{t=1}^n\left(\omega_t x_t^Tu + \sqrt{\omega_t - \omega_t^2}x_t^TW\alpha_t\right)^2\\
& = \frac{d^2}{nm^2}\sum_{t=1}^n\left(\omega_t^2(x_t^Tu)^2 + (\omega_t - \omega_t^2)(x_t^TW\alpha_t)^2 + 2\omega_tx_t^Tu\sqrt{\omega_t - \omega_t^2}x_t^TW\alpha_t\right)\\
& \,{\buildrel d \over =}\, \frac{d^2}{nm^2}\sum_{t=1}^n \left(\omega_t^2b_t^2 + (\omega_t - \omega_t^2)\|x_t^TW\|_2^2 \nu_t + 2 \sigma_t\omega_t x_t^Tu \sqrt{\omega_t - \omega_t^2}\|x_t^TW\|_2 \sqrt{\nu_t}\right)\\
& = \frac{d^2}{nm^2}\sum_{t=1}^n \left(\omega_t^2b_t^2 + (\omega_t - \omega_t^2)c_t^2 \nu_t + 2 \sigma_tb_tc_t \omega_t \sqrt{\omega_t - \omega_t^2}\sqrt{\nu_t}\right)\\
& \triangleq \frac{1}{n}\sum_{t=1}^n B_{1,t} + B_{2,t} + B_{3,t}.
\end{align*}
Here $\nu_t \sim \textrm{Beta}(\frac{1}{2}, \frac{d-2}{2})$, and $\nu_t\equiv1$ if $d=2$, while $\sigma_t$ is a Rademacher random variable, i.e. it takes value $-1$ with probability $\frac{1}{2}$ and value $1$ with probability $\frac{1}{2}$. 
In the last line we define $B_{1,t} \triangleq \frac{d^2}{m^2} \omega_t^2b_t^2$, $B_{2,t} \triangleq \frac{d^2}{m^2}(\omega_t - \omega_t^2)c_t^2\nu_t$ and $B_{3,t} \triangleq \frac{d^2}{m^2}2\sigma_tb_tc_t\omega_t\sqrt{\omega_t - \omega_t^2}\sqrt{\nu_t}$. 

The first equivalence follows from writing $\hat{\Sigma}_1 = \frac{d^2}{nm^2}\sum_{t=1}^n\Phi_t x_t x_t^T \Phi_t$ and grouping the projections instead with the $u$ vectors. 
The second equivalence is just an expansion of the squared term.
For the third equivalence, notice that $x_t^TW \in \RR^{d-1}$ while $\alpha_t \in \RR^{d-1}$ is distributed uniformly on the unit sphere. 
We can think of $\alpha_t$ as a one-dimensional projection operator, and by Fact~\ref{fact:beta} we know that the squared norm of the projection is distributed as a $\textrm{Beta}(\frac{1}{2}, \frac{d-2}{2})$ random variable, scaled by the squared-length of the original vector $x_t^TW$. 
We use the same argument for the third term, except we introduce the Rademacher random variable because the $x_t^TW\alpha_t$ is symmetric about zero. 
The fourth equivalence follows from the fact that $WW^T = I-uu^T$ and therefore $\|x_t^TW\|_2^2 = x_t^TWW^Tx_t = \|x_t\|_2^2 - (x_t^Tu)^2$. 

Now consider all of the $\nu_t$ random variables fixed, and we will develop a deviation bound for the remaining randomness. 
We will apply Bernstein's inequality, so we need to bound the variance and the range.
By Fact~\ref{fact:beta} we know that
\begin{align*}
\Var(B_{1,t}) = \frac{d^4b_t^4}{m^4} \left(\EE \omega_t^4 - (\EE\omega_t^2)^2\right)  = \frac{d^4b_t^4}{m^4}\left( \frac{m(m+2)(m+4)(m+6)}{d(d+2)(d+4)(d+6)} - \left( \frac{m(m+2)}{d(d+2)}\right)^2\right) & \triangleq V_{1,t}\\
\Var(B_{2,t})   = \frac{d^4c_t^4}{m^4} \nu_t^2\left(\frac{m(m+2)}{d(d+2)}\left(1 + \frac{(m+4)(m+6)}{(d+4)(d+6)} - 2\frac{(m+4)}{(d+4)}\right) - \left(\frac{m(d-m)}{d(d+2)}\right)^2\right) &\triangleq V_{2,t}\\
\Var(B_{3,t}) \le \EE (B_{3t}^2) = 4\frac{d^4b_t^2c_t^2}{m^4}\nu_t \frac{m(m+2)(m+4)}{d(d+2)(d+4)}\left(1 - \frac{m+6}{d+6}\right) &\triangleq V_{3,t}
\end{align*}
Therefore,
\begin{align*}
\Var\left(B_{1,t} + B_{2,t} + B_{3,t}\right) \le 3(V_{1,t}+V_{2,t} + V_{3,t}).
\end{align*}
As for the range, by straightforward calculation, we have
\begin{align*}
\left|B_{1,t} + B_{2,t} + B_{3,t}\right| \le \frac{d^2}{m^2} \left(b_t^2 + \frac{1}{4}c_t^2\nu_t + \frac{2}{3}b_tc_t\sqrt{\nu_t}\right).
\end{align*}
The last term is actually maximized when $\omega_t = 3/4$ and takes value $3\sqrt{3}/8 \le 2/3$. 
Bernstein's inequality now reveals that, with probability at least $1-\delta$,
\begin{align*}
\left|u^T\hat{\Sigma}_1 u  - \EE_{\omega_t, \sigma_t} u^T\hat{\Sigma}_1 u\right| & \le \sqrt{\frac{6 \log(2/\delta)}{n}} \times\sqrt{\frac{1}{n}\sum_{t=1}^n (V_{1,t}+V_{2,t}+V_{3,t})} 
\ifthenelse{\equal{\version}{arxiv}}{\\&}{}
+ \frac{2d^2\log(2/\delta)}{3nm^2}\max_{t \in [n]}\left(b_t^2 + \frac{1}{4}c_t^2\nu_t + \frac{2}{3}b_tc_t\sqrt{\nu_t}\right).
\end{align*}
Next we obtain a bound on $\left| \EE_{\omega_t, \sigma_t} u^T\hat{\Sigma}_1 u - u^T\bar{\Sigma}u\right|$, and combine these via the triangle inequality to prove the needed result.
The expectation here is
\begin{align*}
\EE_{\omega_t, \sigma_t} u^T\hat{\Sigma}_1 u = \frac{1}{n}\sum_{t=1}^n\frac{d(m+2)}{m(d+2)}b_t^2 + \frac{d(d-m)}{m(d+2)}c_t^2\nu_t.
\end{align*}
The $B_{3,t}$ is zero in expectation due to the symmetric Rademacher random variable.
So in expectation over $\omega_t, \sigma_t$, by substituting in for $u^T\bar{\Sigma}u$, the left hand side of the application of Bernstein's inequality is
\begin{align}
\left| \EE_{\omega_t, \sigma_t} u^T\hat{\Sigma}_1 u^T - u^T\bar{\Sigma}u\right| = \left|\frac{d(d-m)}{m(d+2)} \left(\frac{1}{n}\sum_{t=1}^n c_t^2\nu_t - \frac{1}{(d-1)}c_t^2\right)\right| = \frac{d(d-m)}{m(d+2)}\left|\frac{1}{n}\sum_{t=1}^n c_t^2\left(\nu_t - \frac{1}{d-1}\right)\right|.
\label{eq:quadratic_expected_deviation}
\end{align}
Note that if $d=2$, this quantity is identically zero.
We are left to control all of the terms involving the $\nu_t$ random variables for $d\geq3$. 
By Proposition~\ref{prop:beta_dev_1} with $s = 1/2$, we have that for any $\delta_1 > 0$, provided that $\log(1/\delta_1) \le \frac{n}{3200}\frac{(d-2)(d+3)^2}{(d-1)^2(d+1)}$,
\begin{align*}
\PP\left( \frac{1}{n}\sum_{t=1}^n c_t^2\left(\nu_t  - \frac{1}{d-1}\right) > \sqrt{\frac{8 (d-2)c^4 \log(1/\delta_1)}{n(d-1)^2(d+1)}}\right) \le \delta_1,
\end{align*}
where $c = \max_{t \in [n]} c_t$. 

By Proposition~\ref{prop:beta_dev_2}, we have that for any $\delta > 0$,
\begin{align*}
\PP\left( \frac{1}{n}\sum_{t=1}^n c_t^2 \left( \frac{1}{d-1}  - \nu_t\right) > \sqrt{\frac{6c^4 \log(1/\delta)}{(d^2-1)n}}\right) \le \delta.
\end{align*}
These two bounds control the upper and lower tails of the right hand side of Equation~\eqref{eq:quadratic_expected_deviation}.
Finally, by Proposition~\ref{prop:beta_dev_3} we have that for any $\delta > 0$ and $d\geq4$,
\begin{align*}
\PP\left(\max_{t \in [n]} \nu_t > \frac{8}{d}\log(n/\delta)\right) \le \delta,
\end{align*}
which is also trivially true for $d=3$ and $d=2$.
This last bound allows us to control the $V_{2,t}, V_{3,t}$ terms.
Specifically, we have
\begin{align*}
V_{1,t} \le \frac{d^4b^4}{m^4}\left( \frac{m(m+2)(m+4)(m+6)}{d(d+2)(d+4)(d+6)} - \left( \frac{m(m+2)}{d(d+2)}\right)^2\right) & \triangleq V'_1\\
V_{2,t} \le \frac{d^4c^4}{m^4} \frac{64\log^2(n/\delta)}{d^2}\left(\frac{m(m+2)}{d(d+2)}\left(1 + \frac{(m+4)(m+6)}{(d+4)(d+6)} - 2\frac{(m+4)}{(d+4)}\right) - \left(\frac{m(d-m)}{d(d+2)}\right)^2\right) & \triangleq V'_{2}\\
V_{3,t} \le \frac{4d^4b^2c^2}{m^4}\frac{8\log(n/\delta)}{d} \frac{m(m+2)(m+4)}{d(d+2)(d+4)}\left(1 - \frac{m+6}{d+6}\right) & \triangleq V'_{3}
\end{align*}
where $b = \max_{t \in [n]} b_t$. 
It also controls the terms $c_t^2\nu_t$ in the range term of our application of Bernstein's inequality. 
Combining all of the bounds gives
\begin{align*}
\left|u^T\hat{\Sigma}_1u - u^T\bar{\Sigma}u\right| \le & \sqrt{\frac{6\log(2/\delta)}{n}}\sqrt{V_1'+V_2'+V_3'} + \\
& + \frac{2d^2\log(2/\delta)}{3nm^2}\left(b^2 + \frac{1}{4}c^2 \frac{8}{d}\log(n/\delta)+ \frac{2}{3}bc\sqrt{\frac{8}{d} \log(n/\delta)}\right) \\
& + \frac{d(d-m)}{m(d+2)}\left(\sqrt{\frac{8(d-2)c^4\log(1/\delta_1)}{n(d-1)^2(d+1)}} + \sqrt{\frac{6c^4\log(1/\delta)}{(d^2-1)n}}\right).
\end{align*}
The second term on the right hand side can be upper bounded by:
\begin{align*}
\frac{2d^2\log(2/\delta)}{3nm^2}\left(b+2c\sqrt{\frac{\log(n/\delta)}{d}}\right)^2 \le \frac{4d^2\log(2/\delta)}{nm^2} \left(b^2 + \frac{c^2\log(n/\delta)}{d}\right).
\end{align*}
While the third term, by setting $\delta = \delta_1$ can be upper bounded by
\begin{align*}
\frac{(\sqrt{8} + \sqrt{6}) d(d-m)}{m(d+2)}\sqrt{\frac{c^4\log(1/\delta)}{n(d^2-1)}} \le \frac{8 d}{m} \sqrt{\frac{c^4\log(1/\delta)}{nd^2}}.
\end{align*}
For this inequality we use that $d^2-1 \ge 3d^2/4$ when $d \ge 2$. 

We have bounds for $V_1',V_2'$, and $V_3'$:
\begin{align*}
V_1' &\le \frac{b^4 d^4 m(m+2)}{m^4 d(d+2)},\\
V_2' &\le \frac{128 c^4 d^4 \log^2(n/\delta)}{m^4 d^2}\frac{m(m+2)}{d(d+2)},\\
V_3' &\le \frac{32d^4 b^2c^2\log(n/\delta)}{m^4 d}\frac{m(m+2)}{d(d+2)}.
\end{align*}
Here we use the fact that $m\le d$ so terms of the form $\frac{(m+x)}{(d+x)} \le 1$. 
This means that for $\delta \le n/e$ we have:
\begin{align*}
& \sqrt{V_1' + V_2' + V_3'} \le \frac{d^2}{m^2} \sqrt{\frac{m(m+2)}{d(d+2)} \log^2(n/\delta)}\left(b^2 + \frac{16c^2}{d}\right)
 \le \sqrt{\frac{3d^2}{m^2}\log^2(n/\delta)}\left(b^2 + \frac{16c^2}{d}\right) .
\end{align*}
Putting everything together proves the claim.
The condition that $\log(1/\delta) \le \frac{n}{9600} \frac{d(d+3)^2}{(d-1)^2(d+1)}$ in the statement is stronger than the one required by Proposition~\ref{prop:beta_dev_1} since we only apply it when $d \ge 3$. 

\subsection{Proof of Lemma~\ref{lem:spectral_tail_conditions}}
\label{sec:spectral_tail_proof}
We first derive the bound for $Y_1$, which is the simplest of the three.
Notice that $\EE Y_1 = (\EE\omega^2) xx^T$ and
\begin{align*}
\EE (Y_1 - \EE Y_1)^p = \EE\left(\omega^2 - \EE\omega^2\right)^p \|x\|^{2(p-1)} xx^T.
\end{align*}
We now proceed to bound the central moments of the random variable $\omega^2$. 
Notice that since $\omega \sim \textrm{Beta}(\tfrac{m}{2}, \tfrac{d-m}{2})$, we know the non-central moments by Fact~\ref{fact:beta}. 
We also have the bound
\begin{align*}
\frac{m+2i}{d+2i} \le 2 \frac{m}{d}(i+1) \qquad \textrm{for } i \ge 0.
\end{align*}
This can be seen by,
\begin{align*}
\frac{d}{m}\frac{m+2i}{d+2i} = \frac{1+2i/m}{1+2i/d} \le 1+2i/m \le 2(i+1),
\end{align*}
for $i \ge 0, d \ge 0, m \ge 1$. 

Now to control the central moment of $\omega^2$, we apply Minkowski's inequality to obtain
\begin{align*}
|\EE(\omega^2 - \EE(\omega^2))^p| &\le \EE|\omega^2 - \EE\omega^2|^p \le \left( (\EE \omega^{2p})^{1/p} + \EE\omega^2\right)^p
= \left( \left( \prod_{i=0}^{2p-1} \frac{m+2i}{d+2i}\right)^{1/p} + \frac{m(m+2)}{d(d+2)}\right)^p.
\end{align*}
Now notice that since $m \le d$ the term $\frac{m+2i}{d+2i} \le 1$ for all $i$ but also the expression is monotonically increasing with $i$. 
This means that we can bound
\begin{align*}
\frac{m(m+2)}{d(d+2)} \le \frac{(m+2i)(m+2(i+1))}{(d+2i)(d+2(i+1))}, \qquad \forall\ i \ge 0.
\end{align*}
This bound implies that the second term above is always smaller than the first, which leads to
\begin{align*}
|\EE(\omega^2 - \EE(\omega^2))^p| &\le 2^p\prod_{i=0}^{2p-1} \frac{m+2i}{d+2i}
= 2^p \frac{m(m+2)(m+4)}{d(d+2)(d+4)} \prod_{i=3}^{2p-1} \frac{m+2i}{d+2i}\\
& \le 2^p \frac{m(m+2)(m+4)}{d(d+2)(d+4)} \prod_{i=3}^{p+1}\frac{m+2i}{d+2i}\\
& = 2^p \frac{m(m+2)(m+4)}{d(d+2)(d+4)} \prod_{i=0}^{p-2}\frac{m+6+2i}{d+6+2i}\\
& \le 2^p \frac{m(m+2)(m+4)}{d(d+2)(d+4)} \left(2\frac{m+6}{d+6}\right)^{p-1}(p-1)!\\
& \le \frac{p!}{2} \left(28 \frac{m}{d}\right)^{p-2} 4\times28\times3\times5 \frac{m^4}{d^4}.
\end{align*}
The first inequality is based on the argument above, that the second term in the application of Minkowski's inequality can be dominated by the first term.
The second inequality follows since $p \ge 2$ so $2p-1 \ge p+1$ and the fact that the terms of the form $\frac{m+x}{d+x}$ are at most $1$.
The third inequality follows from the bound derived above on terms of the form $\frac{m+2i}{d+2i}$. 
The last line follows from the fact that $\frac{m+i}{d+i} \le \frac{m}{d}(i+1)$ applied to all terms of that form and the bound $(p-1)! \le p!/2$. 

Putting things together, we have
\begin{align*}
\EE (Y_1 - \EE Y_1)^p &\preceq \frac{p!}{2} \left(28 \frac{m}{d}\right)^{p-2} 1680 \frac{m^4}{d^4} \|x\|_2^{2(p-1)} xx^T
 = \frac{p!}{2} \left(28 \frac{m}{d}\|x\|_2^2\right)^{p-2} 1680 \frac{m^4}{d^4} \|x\|_2^2 xx^T,
\end{align*}
which proves the first claim.

For the claim involving $Y_2$, notice first that for a unit vector $u$, we can exploit orthogonality to write
\begin{align*}
(a uu^T + b I)^p = ((a+b)uu^T + b (I - uu^T))^p = (a+b)^puu^T + b^p (I - uu^T).
\end{align*}
We will apply this identity on the term involving $Y_2$.
Notice also that,
\begin{align*}
\EE Y_2 = (\EE \omega - \omega^2) \|x\|_2^2 \frac{WW^T}{d-1}.
\end{align*}
Since $W^TW = I_{d-1}$ and since $\alpha$ and $\omega$ are independent and $\|\alpha\|_2 = 1$,
\begin{align*}
\EE (Y_2 - \EE Y_2)^p &= \EE\left( (\omega - \omega^2)\|x\|^2W\alpha \alpha^TW^T - \EE(\omega - \omega^2)\|x\|^2 \frac{WW^T}{d-1}\right)^p\\
& = \|x\|_2^{2p}W \EE\left( (\omega - \omega^2)\alpha\alpha^T - \EE(\omega - \omega^2)\frac{I_{d-1}}{d-1}\right)^pW^T\\
& = \|x\|_2^{2p}W \EE\left[ \left(\omega - \omega^2 - \frac{\EE(\omega - \omega^2)}{d-1}\right)^p\alpha \alpha^T + \left(\frac{-\EE(\omega - \omega^2)}{d-1}\right)^{p}(I_{d-1} - \alpha\alpha^T)\right]W^T\\
& = \|x\|_2^{2p}W \EE\left[ \left(\omega - \omega^2 - \frac{\EE(\omega - \omega^2)}{d-1}\right)^p\frac{I_{d-1}}{d-1} + \left(\frac{-\EE(\omega - \omega^2)}{d-1}\right)^{p}\left(1 - \frac{1}{d-1}\right)I_{d-1}\right]W^T\\
& = \|x\|_2^{2p}WW^T \EE\left[ \left(\omega - \omega^2 - \frac{\EE(\omega - \omega^2)}{d-1}\right)^p\frac{1}{d-1} + \left(\frac{-\EE(\omega - \omega^2)}{d-1}\right)^{p}\left(1 - \frac{1}{d-1}\right)\right].
\end{align*}
As before, we now use Minkowski's inequality to bound the term involving the Beta random variables.
\begin{align*}
& \left|\EE\left[ \left(\omega - \omega^2 - \frac{\EE(\omega - \omega^2)}{d-1}\right)^p\frac{1}{d-1} + \left(\frac{-\EE(\omega - \omega^2)}{d-1}\right)^{p}\left(1 - \frac{1}{d-1}\right)\right]\right|\\
& \le \left|\frac{1}{d-1} \EE \left[ \left(\omega - \omega^2 - \frac{\EE(\omega - \omega^2)}{d-1}\right)^p\right]\right| + \left(\frac{\EE(\omega - \omega^2)}{d-1}\right)^{p}\\
& \le \frac{1}{d-1} \left[\left(\EE(\omega - \omega^2)^p\right)^{1/p} + \frac{\EE(\omega - \omega^2)}{d-1}\right]^{p} + \left(\frac{\EE(\omega - \omega^2)}{d-1}\right)^{p}\\
& \le \frac{1}{d-1} \left[\left(\EE(\omega^p)\right)^{1/p} + \frac{\EE\omega}{d-1}\right]^{p} + \left(\frac{\EE\omega}{d-1}\right)^{p}.
\end{align*}
Here the second line is based on the triangle inequality, while the third line follows from Minkowski's inequality on the first term.
In the fourth line, we use the bound $\EE(\omega - \omega^2) \le \EE(\omega)$ on all terms, which is valid since $\omega \in [0,1]$.
As in the bound for $Y_1$, we now use the fact that
\begin{align*}
\frac{m}{d} \le \frac{m+i}{d+i} \ \forall \ i \ge 0 \Rightarrow \frac{m}{d} \le \left(\prod_{i=0}^{p-1} \frac{m+2i}{d+2i}\right)^{1/p}.
\end{align*}
Applying this bound to both terms involving $\EE\omega = \frac{m}{d}$ gives
\begin{align*}
 \frac{1}{d-1}\left[\left(\EE(\omega^p)\right)^{1/p} + \frac{\EE\omega}{d-1}\right]^{p} + \left(\frac{\EE\omega}{d-1}\right)^{p}
& = \frac{1}{d-1}\left[ \left(\prod_{i=0}^{p-1} \frac{m+2i}{d+2i}\right)^{1/p} + \frac{m}{d(d-1)}\right]^p + \left(\frac{m}{d(d-1)}\right)^p\\
& \le \frac{1}{d-1}\left( (1+\frac{1}{d-1})^p + \frac{1}{d-1}^p\right) \prod_{i=0}^{p-1}\frac{m+2i}{d+2i}\\
& \le \frac{2^{p}+1}{d-1}\prod_{i=0}^{p-1}\frac{m+2i}{d+2i}.
\end{align*}
We now use the same upper bounds as we did to control $Y_1$,
\begin{align*}
\frac{2^{p}+1}{d-1}\prod_{i=0}^{p-1}\frac{m+2i}{d+2i} &= \frac{2^p+1}{d-1}\frac{m(m+2)}{d(d+2)} \prod_{i=1}^{p-2}\frac{m+2+2i}{d+2+2i}\\
& \le \frac{2^{p}+1}{d-1} \frac{m(m+2)}{d(d+2)} \left(2 \frac{m+2}{d+2}\right)^{p-2} (p-1)!\\
& \le \frac{p!}{2}\left(16 \frac{m}{d}\right)^{p-2} 32 \frac{m^2}{d^3}
\end{align*}
Combining this with the derivation above gives:
\begin{align*}
\EE(Y_2 - \EE Y_2)^p \preceq \frac{p!}{2}\left(16 \frac{m}{d}\right)^{p-2} 32 \frac{m^2}{d^3} \|x\|_2^{2p}WW^T
= \frac{p!}{2}\left(16 \frac{m}{d}\|x\|_2^2\right)^{p-2} 32 \frac{m^2}{d^3} \|x\|_2^{4}\left(I - \frac{xx^T}{\|x\|_2^2}\right),
\end{align*}
which, along with the fact that $I - uu^T \preceq I$, gives the bound for $Y_2$.

Finally for $Y_3$, note that $\EE Y_3 = 0$ which is clear because $W$ is an orthonormal basis for the subspace orthogonal to $x$. 
Thus, for odd $p$, we have $\EE Y_3^p = 0$ while for even $p$
\begin{align*}
\EE Y_3^p &= \EE\left( \omega \sqrt{\omega - \omega^2}\right)^p \|x\|_2^p \EE (x\alpha^TW^T + W\alpha x^T)^p\\
& = \EE\left( \omega \sqrt{\omega - \omega^2}\right)^p \|x\|_2^p \left(\|x\|_2^{p-2}xx^T + \|x\|_2^p \frac{WW^T}{d-1}\right)\\
& = \EE\left( \omega \sqrt{\omega - \omega^2}\right)^p \|x\|_2^{2(p-2)} \left(\|x\|_2^2 xx^T + \|x\|^4_2 \frac{WW^T}{d-1}\right).
\end{align*}
The only non-trivial step here is the second one, where we note that $(x\alpha^TW^T + W\alpha x^T)^2 = (xx^T + \|x\|^2 W\alpha\alpha^TW^T)$ by direct calculation and exploiting orthogonality of $x$ and $W$.
By further exploiting orthogonality, this expression to any natural power is equal to taking each term to that power and this gives the expression in the second line. 

Now for the term involving $\omega$, since $\omega \in [0,1]$ and $p$ is even,
\begin{align*}
\EE(\omega \sqrt{\omega - \omega^2})^p &= \EE (\omega^{3/2}\sqrt{1-\omega})^p \le \EE\omega^{3p/2} = \prod_{i=0}^{3p/2-1}\frac{m+2i}{d+2i}\\ 
& = \frac{m(m+2)(m+4)}{d(d+2)(d+4)} \prod_{i=3}^{3p/2-1}\frac{m+2i}{d+2i}\\
& \le 15 \frac{m^3}{d^3} \prod_{i=1}^{3p/2-3}\frac{m+4+2i}{d+4+2i} \le 15 \frac{m^3}{d^3} \prod_{i=1}^{p-2}\frac{m+4+2i}{d+4+2i} \\
& \le 15 \frac{m^3}{d^3} \left(2 \frac{m+4}{d+4}\right)^{p-2}(p-1)! \le \frac{p!}{2}\left(10 \frac{m}{d}\right)^{p-2}15 \frac{m^3}{d^3}.
\end{align*}
This derivation uses all of the same steps as in the previous two cases.
The only thing to note is that we use the bound $3p/2 - 3 \ge p-2$ which holds as long as $p \ge 2$. 
Combining this with above gives
\begin{align*}
\EE Y_3^p \preceq \frac{p!}{2}\left(10 \frac{m}{d}\|x\|_2^2\right)^{p-2}15 \frac{m^3}{d^3}\|x\|_2^2\left(\|x\|^2_2 \frac{2 WW^T}{d}+ xx^T\right).
\end{align*}
The last step is to use the fact that $WW^T = I - \frac{xx^T}{\|x\|_2^2}$.
This proves the lemma. 

\subsection{Proof of Lemma~\ref{lem:kl_bound}}
Let $\Sigma_0 = \eta I, \Sigma_1 = \eta I - \gamma e_1e_1^T$. 
We will prove Lemma~\ref{lem:kl_bound} for the distributions based on $\Ncal(0, \Sigma_0)$ and $\Ncal(0, \Sigma_1)$.
By rotational invariance, the bound holds if we replace $e_1$ with any unit vector $v$. 
The KL-divergence for a single sample is:
\begin{align*}
KL(\PP_1 || \PP_0) &= \int \Ncal(0, U^T\Sigma_1 U) \Unif(U) \log \left( \frac{\Ncal(0, U^T\Sigma_1 U) \Unif(U)}{\Ncal(0, U^T\Sigma_0 U) \Unif(U)}\right)\\
& = \EE_{U \sim \Unif} KL(\Ncal(0, U^T\Sigma_1U) || \Ncal(0, U^T\Sigma_0 U))\\
& = \EE_{U \sim \Unif} \frac{1}{2}\left( \frac{1}{\eta} \tr(\eta I_m - \gamma U^Te_1e_1^TU) - m - \log \frac{\det(\eta I_m - \gamma U^Te_1e_1^TU)}{\eta^m} \right)
\end{align*}
Here $\Unif$ is the density for the uniform distribution over orthonormal bases for $m$-dimensional subspaces of $\RR^d$. 
To analyze the quantity inside the expectation, let $\lambda_1, \ldots, \lambda_m$ denote the eigenvalues of $\eta I_m - \gamma U^Te_1e_1^TU$ and observe that
these eigenvalues are either $\eta$ or $\eta-\gamma$ due to spherical symmetry. 
We have
\begin{align*}
& \frac{1}{2}\left( \frac{1}{\eta}\tr(\eta I_m - \gamma U^Te_1e_1^TU) - m - \log \frac{\det(\eta I_m - \gamma U^Te_1e_1^TU)}{\eta^m} \right)\\
& = \frac{1}{2}\left( \sum_{i=1}^m \lambda_i/\eta - \log (\lambda_i/\eta) - 1\right) 
  \le \frac{1}{2}\sum_{i=1}^m (\lambda_i/\eta -1)^2\\
  & = \frac{1}{2\eta^2} \|U^T(\Sigma_0 - \Sigma_1)U\|_F^2
 = \frac{\gamma^2}{2\eta^2}\|U^Te_1e_1^TU\|_F^2.
\end{align*}
The inequality here is $x - \log(x) - 1 \le (x-1)^2$ which holds for $x \ge 1/e$.
This leads to the condition that $\lambda_i/\eta \ge 1/e$, and since $\lambda_i$ is either $\eta$ or $\eta-\gamma$, we require $\gamma \le \frac{1-e}{e}\eta$ which is the condition in the statement. 
So we have shown
\begin{align*}
KL(\PP_1 || \PP_0) \le \frac{\gamma^2}{2\eta^2}\EE_{U \sim \Unif} \|U^Te_1 e_1^TU\|_F^2.
\end{align*}
We will now upper bound this expectation.
\begin{align*}
\EE_{U \sim \Unif} \|U^Te_1e_1^TU\|_F^2 = \sum_{i,j=1}^m \EE_{U \sim \Unif}U_{1i}^2 U_{1j}^2.
\end{align*}
This is the squared-Frobenius norm of the outer product of the first \emph{row} (in $\RR^m$) with itself. 
Marginally, each entry of $U$, after squaring, is distributed as $\Beta(\frac{1}{2}, \frac{d-1}{2})$ so the diagonal terms of this matrix (the terms where $i=j$ above) are just the second (non-central) moment of the $\Beta$ distribution.
These are $\frac{3}{d(d+2)} \le \frac{3}{d^2}$.

For the off-diagonal terms, note that by spherical symmetry each row of $U$ has a direction that is chosen uniform at random (in $m$ dimensions) while the squared-norm of each row is distributed as $\Beta(\frac{m}{2}, \frac{d-m}{2})$.
This second fact holds because $UU^T e_1 \equalsd \omega e_1 + \sqrt{\omega(1-\omega)}z$ where $z \perp e_1$ and $\omega \sim \Beta(\frac{m}{2}, \frac{d-m}{2})$, so that $e_1^T UU^T e_1 = \|U^Te_1\|_2^2 \equalsd \omega$. 
If we let $v \in \RR^m$ denote a uniform at random unit vector, the off-diagonal terms can be written as
\begin{align*}
\EE \omega^2 v_i^2 v_j^2 = \EE v_i^2 v_j^2 \EE \omega^2 \le \sqrt{\EE v_i^4} \sqrt{\EE v_j^4} \frac{m/2}{d/2}\frac{m/2+1}{d/2+1} = \frac{3}{m(m+2)} \frac{m}{d} \frac{m+2}{d+2} \le \frac{3}{d^2}.
\end{align*}
Here we use the Cauchy-Schwarz inequality, the fact that $\omega \sim \Beta(\frac{m}{2}, \frac{d-m}{2})$ and that marginally each $v^2_j \sim \Beta(\frac{1}{2}, \frac{m-1}{2})$ since $v$ is a uniform random vector in $m$-dimensions. 
So every term in the sum is bounded by $3/d^2$.
There are $m^2$ terms producing the bound
\begin{align*}
\EE_{U \sim \Unif} \|U^Te_1 e_1^TU\|_F^2 \le \frac{3m^2}{d^2}.
\end{align*}
Plugging into our KL bound above and using additivity of KL-divergence for product measures completes the proof.

\subsection{Proof of Lemma~\ref{lem:packing}}
The proof is based on the probabilistic method. 
We will first show that for any fixed unit vector $x$, if we draw another vector $v$ uniformly at random, then
\begin{align*}
\PP[\|xx^T - vv^T\|_2 \le \tau] \le \exp\left( \frac{-d\log(1/\tau) + 1}{4}\right).
\end{align*}
Equipped with this deviation bound, if we draw $M$ points uniformly at random from the unit sphere in $d$ dimensions, then the probability that no two points are within $\tau$ of each other is (via a union bound)
\begin{align*}
\PP[\forall i\ne j: \|v_i v_i^T - v_j v_j^T\|_2 \ge \tau] = 1 - \PP\left[\bigcup_{i\ne j} \|v_i v_i^T - v_j v_j^T\|_2 \le \tau\right] \ge 1 - {M \choose 2} \exp\left( \frac{-d\log(1/\tau) + 1}{4}\right).
\end{align*}
As long as this probability is non-zero, then we know that there exists such a packing set. 
In particular, if:
\begin{align*}
\frac{M(M-1)}{2}\exp\left( \frac{-d\log(1/\tau) + 1}{4}\right) \le 17/18,
\end{align*}
then we would show existence of a packing set of size $M$.  To
proceed, set $\tau=1/2$ and assume that $d \ge 4$, which is a
pre-condition for the lemma. We now proceed to verify that inequality
above is satisfied with $M \leq \exp(d/8\cdot\log(2) - 1/8) + 1$. With
this choice, the inequality above reduces to
\begin{align*}
\exp(d/8\cdot \log(2) - 1/8) \leq 8/9 \exp(d/4\cdot\log(2) - 1/4),
\end{align*}
and with further calculations, we arrive at the condition
\begin{align*}
\frac{d}{8}\log(2) \geq \frac{1}{8} + \log(9/8).
\end{align*}
This latter condition is satisfied with $d\ge 4$, and so we may take
$M = \exp(-1/8)2^{d/8}$, which yields the result.

We prove the deviation bound more generally for $\tau \in
[0.5,1]$. Note that since $x,v$ are unit vectors, the spectral norm
difference is just the magnitude of the sine of the angle between the
two vectors.  It is well known that
(see~\cite{dasgupta2003elementary})
\begin{align*}
\PP\left[ (x^Tv)^2 \ge \beta/d\right] &\le \exp\left(\frac{1}{2}(1-\beta + \log \beta)\right),\\
\PP\left[ \cos^2(\angle(x,v)) \ge (1+\epsilon)/d \right] &\le \exp\left(\frac{1}{2}(-\epsilon + \log (1+\epsilon))\right).
\end{align*}
Therefore,
\begin{align*}
\PP\left[|\sin\angle(x,v)| \le \sqrt{1-(1+\epsilon)/d}\right] = \PP\left[ \cos^2\angle(x,v) \ge (1+\epsilon)/d\right] \le \exp\left( \frac{1}{2}(-\epsilon + \log(1+\epsilon))\right).
\end{align*}
If $\epsilon \ge 3$, then $\epsilon - \log(1+\epsilon) \ge \epsilon/2$, so we can upper bound the probability by $\exp\left(-\epsilon/4\right)$. 
Setting $\epsilon = d(1-\tau^2)-1$ gives the inequality
\begin{align*}
\PP\left[|\sin\angle(x,v)| \le \tau \right] \le \exp\left( \frac{-d(1-\tau^2) + 1}{4}\right).
\end{align*}
We now proceed to lower bound $(1-\tau^2)$ by $\log(1/\tau)$.
This is possible for $\tau \in [0.5,1]$ as both functions are monotonically decreasing in $\tau$ but $(1-\tau^2)$ is concave while $\log(1/\tau)$ is convex. 
At $\tau = 1/2$ the first is larger than the second, and they are both equal at $\tau = 1$.
The condition on $\tau$ and this lower bound establishes the inequality used above.

\subsection{Proof of Lemma~\ref{lem:compressed_vector_estimation}}
\label{sec:compressed_vector_estimation_pf}
Recall that the quantity we are interested in lower bounding is
\begin{align*}
\inf_{T} \EE_{x, \Pi, T} \|T(\Pi, \Pi x) - x\|_2^2.
\end{align*}
The expectation over $T$ allows for randomized estimators, $x$ is drawn uniformly at random from the $\nu$-radius sphere, and $\Pi$ is a uniformly drawn $m$ dimensional projection operator. 
Instead of drawing an $m$-dimensional projection matrix $\Pi$ uniformly at random, it is equivalent to draw an orthonormal basis $U \in \RR^{d \times m}$ uniformly at random.
The observation is then $(U, U^Tx)$ which is clearly equivalent to observing $(\Pi, \Pi x)$ since one can be constructed from the other.
So we will instead lower bound
\begin{align*}
\inf_{T} \EE_{x, U, T} \|T(U, U^Tx) - x\|_2^2 = \inf_{T} \EE_{x,U,T} \|T(U, U^Tx)\|_2^2 - 2T(U, U^Tx)^Tx + \nu^2.
\end{align*}
Before proceeding, we need to clarify one definition.
We will use $\angle V,y=\angle (P_V y, y)$ to denote the angle between the subspace $V$ and the vector $y$.
We can evaluate the integrals by first choosing a subspace $U$, choosing a vector $y \in U$, and finally choosing the vector $x$ so that $P_Ux = y$.
This gives
\begin{align*}
\inf_{T} \EE_{x, \Pi, T} \|T(\Pi, \Pi x) - x\|_2^2 & = \inf_T \int_{U} \int_{y} \int_{x} \left[\EE_T \|T(U, U^Tx)\|_2^2 - 2T(U, U^Tx)^Tx + \nu^2\right] dP(x;y,U)dQ(y)dR(U),
\end{align*}
where $P$ is the conditional distribution of $x$ given that it projects to $y$ with subspace $U$, $Q$ is the distribution over projection vectors $y$, and $R$ is the uniform distribution on the $m$-dimensional Grassmannian manifold.
Now we will push the $\inf_T$ inside of the first two integrals. 
Notice that all of the information the estimator has is $(U,y)$ since $U^Tx = U^Ty$, so for each $(U,y)$ pair, the estimator is just a distribution over vectors.
Calling this distribution $T(\cdot; y,U)$, we can write
\begin{align*}
\inf_{T} \EE_{x, \Pi, T} \|T(\Pi, \Pi x) - x\|_2^2 & \ge \int_U \int_y \inf_{T(\cdot ;y,U)} \int_{v} \int_x \|v\|_2^2 -2v^Tx + \nu^2 dP(x;y,U)dT(v;y,U) dQ(y)dR(U).
\end{align*}
The only term depending on $x$ is the $v^Tx$ term, for which
\begin{align*}
\int_x 2v^TxdP(x;y,U) = 2v^T \EE_{x\sim P(x;y,U)} x = 2v^Ty
\end{align*}
follows by spherical symmetry, since we draw $x$ uniformly from the $\nu$-radius sphere, which is symmetric about the subspace $U$.
Therefore, we have
\begin{align*}
\inf_{T(\cdot ;y,U)}\int_{v} \int_x \left(\|v\|_2^2 -2v^Tx + \nu^2\right) dP(x;y,U)dT(v:y,U) &= \inf_T \int_v \left(\|v\|_2^2 - 2v^Ty + \nu^2\right) dT(v;y,U)\\
& \ge \nu^2 - \|y\|_2^2.
\end{align*}

So we can lower bound by
\begin{align*}
\inf_{T} \EE_{x, \Pi, T} \|T(\Pi, \Pi x) - x\|_2^2 & \ge \nu^2 - \int_U \int_y \|y\|_2^2 dQ(y)dR(U),
\end{align*}
and we are left to control the expected norm of $y = P_Ux$.
We have the identity $\|y\| = \|x\| \cos \angle (P_U x, x) = \nu \cos \angle (P_U x, x)$.
By spherical symmetry, we can let $U$ to be the span of the first $m$ standard basis vectors, in which case
\begin{align*}
\cos \angle P_Ux, x = \frac{\sqrt{\sum_{i=1}^m x_i^2}}{\sqrt{\sum_{i=1}^d x_i^2}}.
\end{align*}
Therefore, $\|y\|_2^2 \,{\buildrel d \over =}\, \nu^2 Z = \nu^2 \frac{\sum_{i=1}^m x_i^2}{\sum_{i=1}^d x_i^2}$ where $x_1, \ldots, x_d \sim \Ncal(0,1)$.
This gives the lower bound
\begin{align*}
\inf_{T} \EE_{x, \Pi, T} \|T(\Pi, \Pi x) - x\|_2^2\ge \nu^2(1- \EE[Z]) = \nu^2(1 - \frac{m}{d}).
\end{align*}

\end{document}